\documentclass[dvipsnames]{article} %
\usepackage{iclr2027_conference,times}

\iclrfinalcopy %

\usepackage{url}
\usepackage{bm}
\usepackage{xspace}
\usepackage{amsfonts}       %
\usepackage{amssymb}
\usepackage{nicefrac}       %
\usepackage{xspace}
\usepackage{multirow}
\usepackage{optidef}
\usepackage{bm}
\usepackage{mathtools}
\usepackage{amsmath,amsthm}

\newcommand{\methodname}{\texttt{A2D}\xspace}

\usepackage{pifont}

\usepackage{amsmath,amsfonts,bm}

\def\eqref#1{equation~\ref{#1}}

\def\1{\bm{1}}

\DeclareMathAlphabet{\mathsfit}{\encodingdefault}{\sfdefault}{m}{sl}
\SetMathAlphabet{\mathsfit}{bold}{\encodingdefault}{\sfdefault}{bx}{n}

\usepackage{lipsum}

\usepackage{siunitx} 
\usepackage{color, colortbl}
\definecolor{Gray}{gray}{0.9}
\definecolor{LightGray}{gray}{0.97}
\newcolumntype{g}{>{\columncolor{LightGray}}S}

\usepackage{bm}

\usepackage{xcolor}  
\usepackage{multicol}
\usepackage{enumitem}

\usepackage{minitoc}

\usepackage{amsfonts}

\usepackage{setspace}

\usepackage[%
  colorlinks = true,
  citecolor  = Blue,
  linkcolor  = Blue,
  urlcolor   = Blue,
  unicode,
  pagebackref
  ]{hyperref}
\usepackage{hyperref}

\usepackage{amssymb}
\usepackage{amsthm}

\usepackage{amsmath}
\usepackage{graphicx}
\usepackage{cleveref}
\usepackage{float}
\usepackage{algorithm}
\usepackage{algorithmicx}
\usepackage[noend]{algpseudocode}
\usepackage{booktabs}
\usepackage{subcaption}  %
\usepackage{nicefrac}
\usepackage{wrapfig}

\theoremstyle{plain}

\theoremstyle{definition}

\theoremstyle{remark}

\Crefname{equation}{Eq.}{Eqs.} 
\Crefname{section}{Sec.}{Secs.}
\Crefname{table}{Tab.}{Tabs.}
\Crefname{algorithm}{Alg.}{Algs.}

\title{Enhancing Diffusion Language Models with Autoregressive Post-Training Weights}

\author{%
  Yiming Qin\thanks{Equal contribution.} \\
  EPFL, Lausanne, Switzerland\\
  \texttt{yiming.qin@epfl.ch}
  \And
  Ke Wang\footnotemark[1] \\
  EPFL, Lausanne, Switzerland\\
  \texttt{k.wang@epfl.ch}
  \And
  Amel Abdelraheem \\
  EPFL, Lausanne, Switzerland\\
  \texttt{amel.abdelraheem@epfl.ch}
  \And
  Adam Hazimeh \\
  EPFL, Lausanne, Switzerland\\
  \texttt{adam.hazimeh@epfl.ch}
  \And
  Pascal Frossard \\
  EPFL, Lausanne, Switzerland\\
  \texttt{pascal.frossard@epfl.ch}
}

\begin{document}

\doparttoc %
\faketableofcontents %

\maketitle

\lhead{Preprint}

\begin{abstract}

Diffusion language models (dLLMs) have emerged as a promising alternative to autoregressive (AR) language models, offering flexible token-update orders and parallel decoding. Recent dLLMs are often initialized from pretrained AR models before diffusion conversion in order to inherit their learned representations. After the conversion, however, they typically ignore the extensive post-training ecosystem of their AR ancestors. In this work, we show that these existing AR post-training weight updates can instead be effectively recycled to enhance diffusion models. Despite the changes by AR-to-diffusion conversion, directly adding an AR post-training weight update to a diffusion base model remains effective, bringing its performance close to that achieved by direct diffusion post-training. Notably, AR and diffusion post-training updates are nearly orthogonal in weight space, yet induce substantially more aligned representation changes in the diffusion model. Their distinct updates are also complementary: composing their weights can retain gains from both regimes and further improve the post-trained diffusion  model.
Based on these findings, we propose \textbf{A2D}, a simple training-free framework for enhancing diffusion models with existing AR post-training resources. A2D can transfer capabilities from AR post-trained models to diffusion base models, and further improve already post-trained diffusion models by composing AR and diffusion post-training updates. Across various dLLMs, including Dream, DreamReasoner, DiffuCoder, Dream-Coder, Nemotron-Labs-Diffusion, and DiffusionGemma, A2D reliably improves instruction following, mathematical reasoning, and coding with both supervised fine-tuning and reinforcement learning updates, without additional training, or inference-time computation.
\looseness=-1

\end{abstract}

\section{Introduction}
\label{sec:intro}

Diffusion large language models (dLLMs) have recently emerged as a competitive
alternative to autoregressive (AR) models, achieving strong performance across
instruction following, code generation, and mathematical reasoning~\citep{zhu2025llada,ye2025dream,gong2025diffucoder,fu2026nemotronlabsdiffusion,diffugemma2026technicalreport}.
By casting sequence generation as iterative denoising rather than fixed
left-to-right factorization, dLLMs support bidirectional context and flexible
token-update orders, as well as parallel decoding, making them naturally amenable to
scalable generation.

Recent dLLMs commonly initialize from pretrained AR models before diffusion conversion~\citep{ye2025dream,fu2026nemotronlabsdiffusion,wu2026dreamreasoner8bblocksizecurriculumlearning,diffugemma2026technicalreport}, reusing language representations learned during large-scale AR pretraining and substantially improving training efficiency over training dLLMs from scratch.
This pipeline, however, creates a notable asymmetry: dLLMs routinely inherit AR pretraining, yet largely discard the extensive AR post-training investment built upon it. This is especially notable as the AR ecosystem~\citep{qwen3,gemmateam2026gemma4,liu2026ministral3} often provides a rich collection of post-trained checkpoints across diverse capabilities, which are often stronger than their diffusion    counterparts. Diffusion models instead typically develop these capabilities through separate
diffusion-specific post-training, 
where learning must span varying corruption
levels and partially observed contexts, posing optimization challenges distinct
from standard AR post-training
\citep{parashar2026learnability, lin2026ar}. \looseness=-1

This raises a natural question:
\emph{can the knowledge acquired with AR post-training be incorporated into
diffusion models?}
Such a transfer could reuse readily available, often stronger AR post-trained checkpoints while retaining
the parallel-decoding benefits of diffusion.
Since AR post-training and diffusion conversion originate from the same pretrained
AR model, their shared initialization and architecture naturally motivate a weight-space approach, namely, \textit{model merging}.
Within the AR ecosystem, model merging has proven effective at integrating
capabilities from independently fine-tuned checkpoints without requiring their training data~\citep{ wortsman2022modelsoup, yang2026model}.
In particular, task arithmetic represents post-training weight updates as \textit{task vectors} that can be transferred or composed across compatible models~\citep{ilharco2022editing}.
Extending model merging across the AR--diffusion boundary, however, is nontrivial: diffusion conversion moves the model away from its AR initialization while changing generation from causal next-token prediction to bidirectional denoising. \looseness=-1

\begin{figure}[t]
    \vspace{-5pt}
    \centering
    \begin{subfigure}[t]{0.45\linewidth}
        \centering
        \includegraphics[width=\linewidth]
        {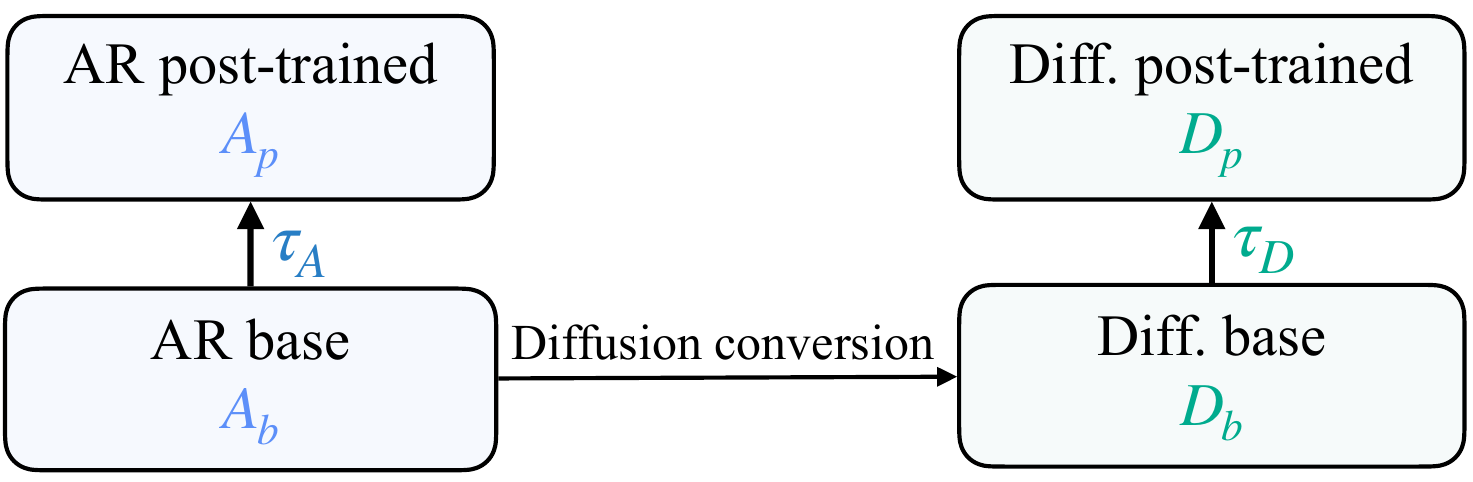}
        \caption{Illustration of notations.}
        \label{fig:main}
    \end{subfigure}
    \hfill
    \begin{subfigure}[t]{0.45\linewidth}
        \centering
        \includegraphics[width=\linewidth]
        {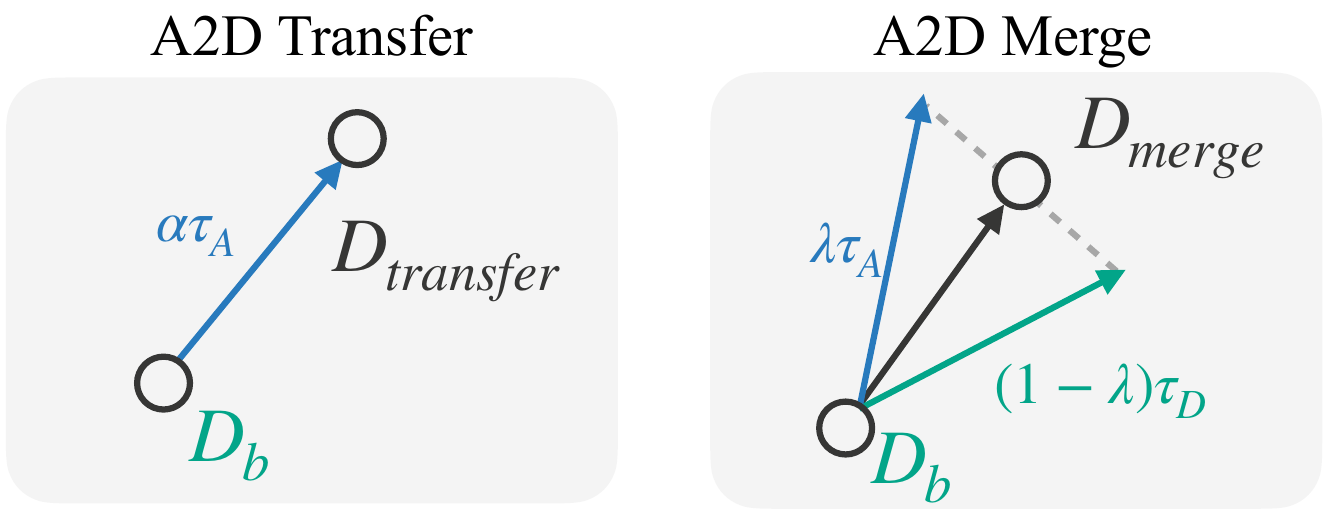}
        \caption{Illustration of \methodname.}
        \label{fig:task_vector}
    \end{subfigure}
    \vspace{-4pt}
    \caption{\textbf{Illustration of notations and method}. {(a)} Diffusion base models with weights $\bm{D}_b$ are often adapted from AR base models with weights $\bm{A}_b$. The two models often undergo separate post-training stages, leading to $\bm{D}_p$ and $\bm{A}_p$, with the corresponding task vectors denoted by $\bm{\tau}_D$ and $\bm{\tau}_A$, respectively. {(b)} \methodname-Transfer directly adds the AR post-training vector to diffusion base $\bm{D}_b$ to create a post-trained diffusion model. \methodname-Merge interpolates this model and the diffusion model's own post-trained checkpoint to further improve downstream capabilities. }
    \label{fig:transfer_and_similarity}
    \vspace{-11pt}

\end{figure}

\begin{wrapfigure}{r}{0.46\textwidth}
\centering
\vspace{-10pt}
\begin{subfigure}[t]{0.49\linewidth}
\centering
\includegraphics[width=\linewidth]{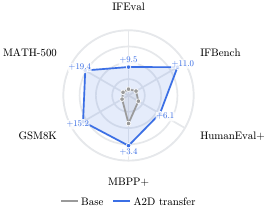}
\caption{Direct transfer}
\label{fig:method_name_radar_base}
\end{subfigure}
\hfill
\begin{subfigure}[t]{0.49\linewidth}
\centering
\includegraphics[width=\linewidth]{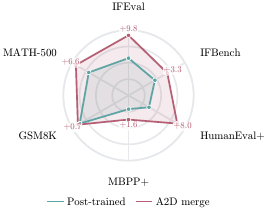}
\caption{Update merge}
\label{fig:method_name_radar_post}
\end{subfigure}
\caption{\textbf{\methodname on Dream.} \methodname improves diffusion base and post-trained models \textbf{without training} through (a) direct AR update transfer and (b) AR--diffusion update merging. 
}
\vspace{-10pt}
\label{fig:example_dream}
\end{wrapfigure}

Nevertheless, we show that AR post-training exhibits \textit{cross-paradigm transferability}:
knowledge learned in the AR regime can remain effective to a
diffusion model without further adaptation. 
In a controlled setting, 
directly transferring the AR post-training task vector brings the
diffusion base close to the performance of direct diffusion post-training. 
This cross-paradigm transfer of AR task vector provides a practical way to improve diffusion
models when no corresponding diffusion post-trained checkpoint is available. 
Moreover, we show that AR and diffusion post-training yield nearly orthogonal task vectors in parameter space, yet induce much more aligned representation changes, suggesting that related capabilities arise through distinct parameter directions.
Motivated by prior findings that less-correlated fine-tuning directions can combine favorably through interpolation~\citep{wortsman2022modelsoup}, we further interpolate the AR-transferred and diffusion-post-trained solutions, yielding performance beyond both endpoints by combining their partially
complementary behavioral gains. \looseness=-1

We introduce \textbf{\methodname} (\emph{AR-to-Diffusion Transfer and Composition}), a simple framework
that exploits the cross-paradigm transferability of AR post-training updates.
\methodname operates in two regimes, as shown in \autoref{fig:task_vector}: (a) in the absence of a diffusion post-training vector, it transfers an AR task vector directly to the diffusion base model; (b) when both
AR and diffusion vectors are available, it interpolates the corresponding
diffusion-space solutions to further improve the post-trained diffusion model.
Both operations are training-free given the available checkpoints and incur no additional trainable parameters.
Across multiple diffusion model families and diverse post-training settings,
including instruction tuning, reasoning, mathematical reasoning, and coding,
both modes of \methodname reliably improves downstream performance, as shown by the example in \autoref{fig:example_dream}.
We summarize our contributions as follows:

\begin{itemize}
    \item We study \emph{cross-paradigm transferability}: AR post-training vectors remain effective after diffusion conversion, inducing aligned representation changes despite near-orthogonal parameter directions, which further motivates their composition with diffusion post-training. 
    \item Building on this observation, we introduce {\methodname}, a simple
    weight-space framework for reusing AR post-training in diffusion models.
    \methodname supports both direct transfer to diffusion base models and training-free
    composition with existing diffusion post-training updates.
    \item Across recent dLLMs including Dream, Dream-Coder, DiffuCoder, Nemotron-Labs-Diffusion, DreamReasoner, and DiffusionGemma, we validate both modes of \methodname and diverse
    post-training scenarios, improving downstream performance without additional training or inference-time computation.
\end{itemize}

\section{Cross-Paradigm Post-Training Transfer and Composition}
\label{sec:motivation}

\subsection{Background and Notations}

\textbf{Autoregressive and diffusion language models.}
AR and diffusion models differ in both their training objectives and generation processes, yet
modern dLLMs are often adapted from pretrained AR checkpoints to inherit the learned language representations
~\citep{ye2025dream,fu2026nemotronlabsdiffusion,wu2026dreamreasoner8bblocksizecurriculumlearning,diffugemma2026technicalreport}.
We denote the weights of this AR source model by $\bm{A}_b$, which is trained with an autoregressive objective and generates tokens from left to right.
Starting from $\bm{A}_b$, further diffusion training adapts the model for diffusion generation
to obtain $\bm{D}_b$.
Depending on the model family, this adaptation uses either a diffusion objective
alone~\citep{ye2025dream} or a combination of AR and diffusion objectives~\citep{fu2026nemotronlabsdiffusion}, and may involve hundreds of billions of tokens of training.
Consequently, despite sharing an initialization, $\bm{D}_b$ can differ substantially
from $\bm{A}_b$ in both parameters and generation behavior. At inference time, it
generates through diffusion-style iterative denoising with flexible token-update orders. 

Starting from their respective base checkpoints, the AR and diffusion models
can be post-trained independently for the same downstream capability, yielding
$\bm{A}_p$ from $\bm{A}_b$ and $\bm{D}_p$ from $\bm{D}_b$, as illustrated in
~\autoref{fig:main}.
Although $\bm{A}_p$ and $\bm{D}_p$ target the same capability, their post-training updates
are learned under different objectives and from different base checkpoints.
This raises our central question: \textit{can an update learned on the AR branch remain
effective when applied to the diffusion branch?} \looseness=-1

\textbf{Model merging and task arithmetic.} Since AR post-training and diffusion conversion originate from the same pretrained model, it naturally motivates weight-space transfer, such as model merging.
Model merging combines knowledge from multiple trained models directly in
weight space without additional training~\citep{yang2026model, wortsman2022modelsoup}.
Task arithmetic represents the effect of post-training as a task vector, defined
as the parameter difference between a post-trained model and its base
~\citep{ilharco2022editing}.
These vectors can be scaled and composed in weight space to transfer or combine
their learned capabilities.
Accordingly, we define the AR and diffusion post-training vectors as
$\bm{\tau}_A=\bm{A}_p-\bm{A}_b$ and $\bm{\tau}_D=\bm{D}_p-\bm{D}_b$, respectively. 
Notably, task arithmetic typically composes task vectors defined relative to a
shared base model or initialization, whereas $\bm{\tau}_A$ and $\bm{\tau}_D$ originate
from different base checkpoints.

\subsection{Cross-Paradigm Transfer of AR Post-Training to Diffusion}
\label{sec:controlled_transfer}
We first examine whether AR post-training can transfer across the AR and diffusion paradigms.
We begin our analysis by testing whether post-training knowledge learned in the AR regime survives
AR-to-diffusion conversion.
In this case, $\bm{\tau}_A$ is learned around $\bm{A}_b$ but applied to its diffusion descendant $\bm{D}_b$, whose parameters and generation mechanism have been substantially reshaped by diffusion conversion.
To isolate the effect of this paradigm shift, we post-train the AR and diffusion base models on identical data for the same downstream capability in a controlled setting. \looseness=-1

\textbf{Experimental setting.}
We consider mathematical reasoning and coding tasks.
For math, we use Qwen2.5-7B and its diffusion counterpart
Dream-7B-Base, obtained from Qwen2.5-7B through AR-to-diffusion
conversion, and fine-tune both on the same 100K NuminaMath examples
~\citep{numina_math_datasets}.
For coding, we use Qwen2.5-Coder-7B and Dream-Coder-v0-Base-7B, its
corresponding diffusion counterpart, and fine-tune both on the same 100K
EpiCoder examples~\citep{wang2025epicoder}.
This construction yields matched AR and diffusion task vectors targeting the
same capability, allowing us to isolate whether AR post-training transfers
across the two generative paradigms.
Full training and evaluation details are provided in
Appendix~\ref{app:motivation_details}.

\paragraph{Cross-paradigm transfer.} We test this directly by applying the scaled AR task vector $\bm{\tau}_A$ to the diffusion
base model:
$
    \bm{D}_{\text{Transfer}}
    =
    \bm{D}_b + \alpha \bm{\tau}_A,
    \ \alpha \geq 0,
    \label{eq:ar_transfer}
$
where $\alpha$ controls the transfer strength. This  modifies only the model parameters while preserving the diffusion inference procedure.

\begin{figure}[t]
    \centering
    \begin{subfigure}[t]{0.48\linewidth}
        \centering
        \includegraphics[width=\linewidth]
        {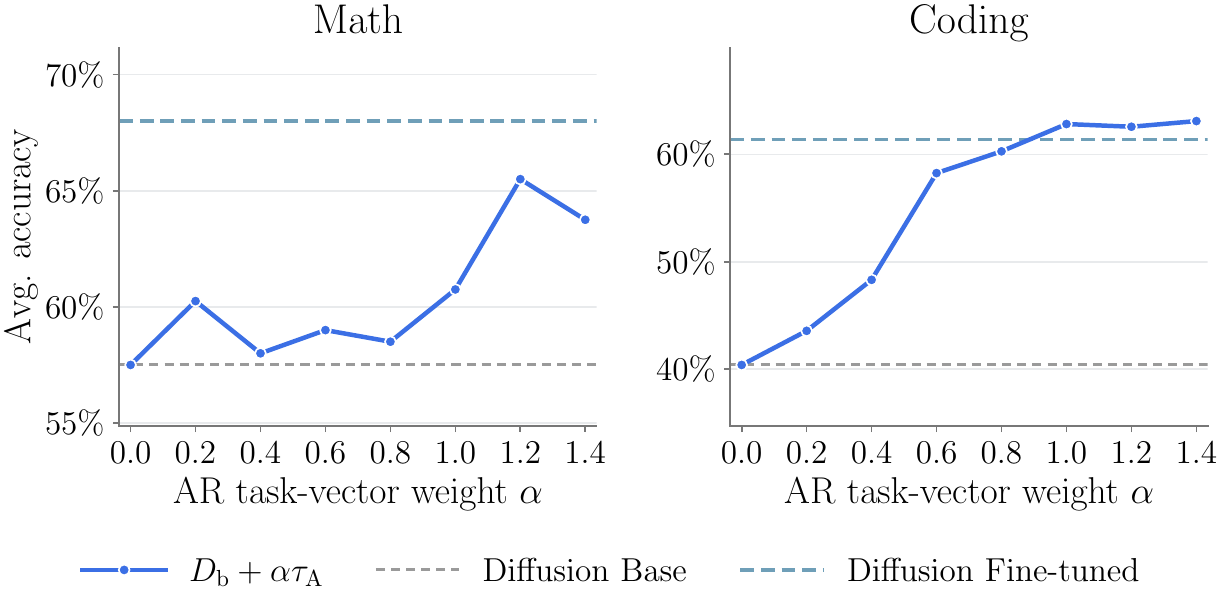}
        \caption{AR post-training updates transfer effectively to diffusion.}
        \label{fig:transfer_sweep}
    \end{subfigure}
    \hfill
    \begin{subfigure}[t]{0.48\linewidth}
        \centering
        \includegraphics[width=\linewidth]
        {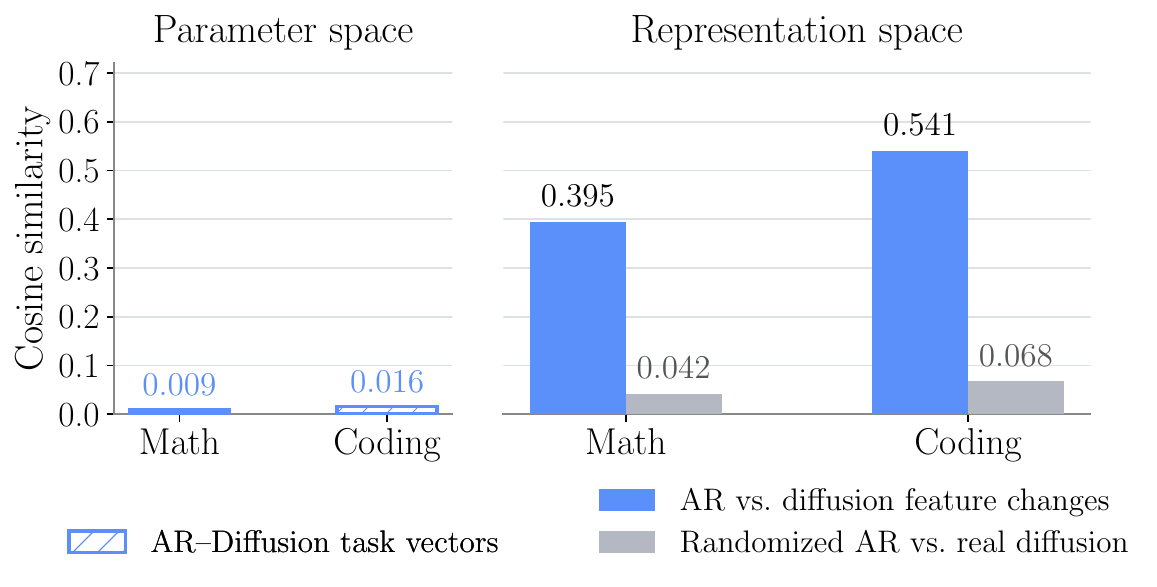}
        \caption{Distinct parameter updates induce aligned representation changes.}
        \label{fig:functional_similarity}
    \end{subfigure}
    \caption{\textbf{Cross-paradigm transfer of AR post-training.} \textbf{(a)} AR task vectors remain effective on the diffusion base. \textbf{(b)} Despite weak parameter alignment, AR and diffusion post-training induce more aligned representation changes than randomized controls.}
    \label{fig:transfer_and_similarity}
\end{figure}

We evaluate mathematical reasoning on GSM8K~\citep{cobbe2021gsm8k} and
MATH-500~\citep{hendrycks2021math} \footnote{We subsample first 200 samples for GSM8K and MATH-500 for the analysis in \autoref{sec:motivation}}, and coding on
HumanEval+~\citep{chen2021codexhumaneval} and
MBPP+~\citep{austin2021mbpp}, respectively for the models fine-tuned on math and coding datasets.
As shown in ~\autoref{fig:transfer_sweep}, transferring $\bm{\tau}_A$ substantially improves $\bm{D}_b$ in both domains and recovers a large fraction of the gains obtained by direct diffusion post-training, or even surpasses them.
This shows that an update learned under causal AR computation can remain effective after
the model has been converted to iterative diffusion denoising. \looseness=-1

\paragraph{Distinct parameters, aligned effects in representation space.}
The successful transfer raises a question: do AR and diffusion
post-training simply learn similar parameter updates?
Despite being trained toward the same capability, $\bm{\tau}_A$ and $\bm{\tau}_D$ are
nearly orthogonal in parameter space with a cosine similarity of only $0.016$ for Coding and $0.009$ for Math.
Parameter-space similarity, however, does not reflect similarity in the induced
model behavior~\citep{xu2024training}.
We therefore compare the representation changes produced by the two task vectors
within the diffusion model.
We generate trajectories with $\bm{D}_b$ and collect intermediate pre-commit masked
states $m$.
For each state, we evaluate $\bm{D}_b$, $\bm{D}_b+\bm{\tau}_A$, and $\bm{D}_b+\bm{\tau}_D$ on the same
input, and isolate the effect of each task vector by subtracting the
representation of the diffusion base:
\[
\Delta \bm{h}_A(m)
=
\bm{h}(\bm{D}_b+\bm{\tau}_A;m)-\bm{h}(\bm{D}_b;m),
\qquad
\Delta \bm{h}_D(m)
=
\bm{h}(\bm{D}_b+\bm{\tau}_D;m)-\bm{h}(\bm{D}_b;m),
\]
where $\bm{h}$ denotes the final-layer representation at the prediction position.
We then measure the cosine similarity between $\Delta \bm{h}_A(m)$ and $\Delta \bm{h}_D(m)$.
As shown in ~\autoref{fig:functional_similarity}, the induced representation
changes are substantially more aligned than the task vectors themselves and
well above randomized-vector controls 
\footnote{The randomized vector is built to share the same norm, global mean and variance as the AR task vector. \looseness=-1}
Together, these results show that AR and diffusion post-training induce related
representation changes, despite following substantially different directions in
parameter space.

\subsection{Composing Autoregressive and Diffusion Post-Training}
\label{sec:composition}

The preceding results reveal an interesting structure: AR and diffusion
post-training induce related representation changes through largely distinct
parameter directions.
Prior work has suggested that less-correlated fine-tuning directions can admit
favorable linear interpolations~\citep{wortsman2022modelsoup}, motivating the composition of these two post-training models.
We therefore examine whether their composition can yield a stronger model than either alone.

To analyze this composition in weight space, we consider the two-dimensional task-vector plane
$
    \bm{D}(\alpha,\beta)
    =
    \bm{D}_b+\alpha\bm{\tau}_A+\beta\bm{\tau}_D,
    \ \alpha,\beta\geq0,
    \label{eq:2d_composition}
$
where $\alpha$ and $\beta$ controls the contribution of the AR and diffusion task vector, respectively.
The AR-transferred and diffusion-post-trained models correspond to
$\bm{D}_T=\bm{D}(1,0)$ and $\bm{D}_p=\bm{D}(0,1)$, respectively, while
$\alpha+\beta=1$ traces the linear interpolation between them. \looseness=-1

\begin{figure}[t]
    \centering
    \begin{subfigure}[t]{0.59\linewidth}
        \centering
        \includegraphics[width=\linewidth]
        {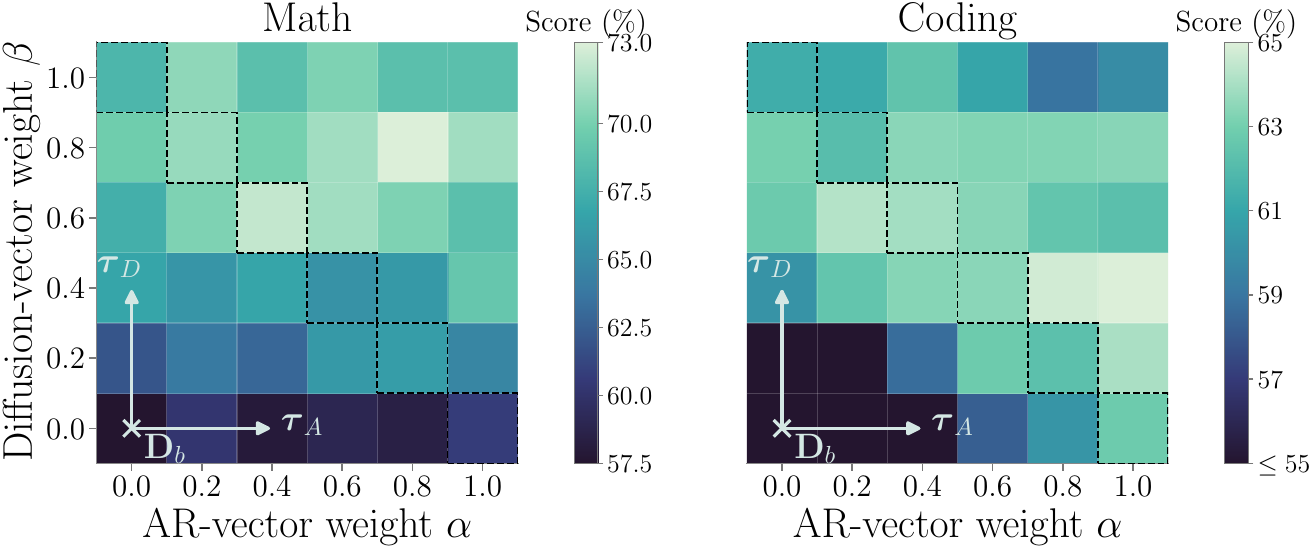}
        \caption{High-performing solutions lie between the AR and diffusion endpoints.}
        \label{fig:composition_landscape}
    \end{subfigure}
    \hfill
    \begin{subfigure}[t]{0.35\linewidth}
        \centering
        \includegraphics[width=\linewidth]{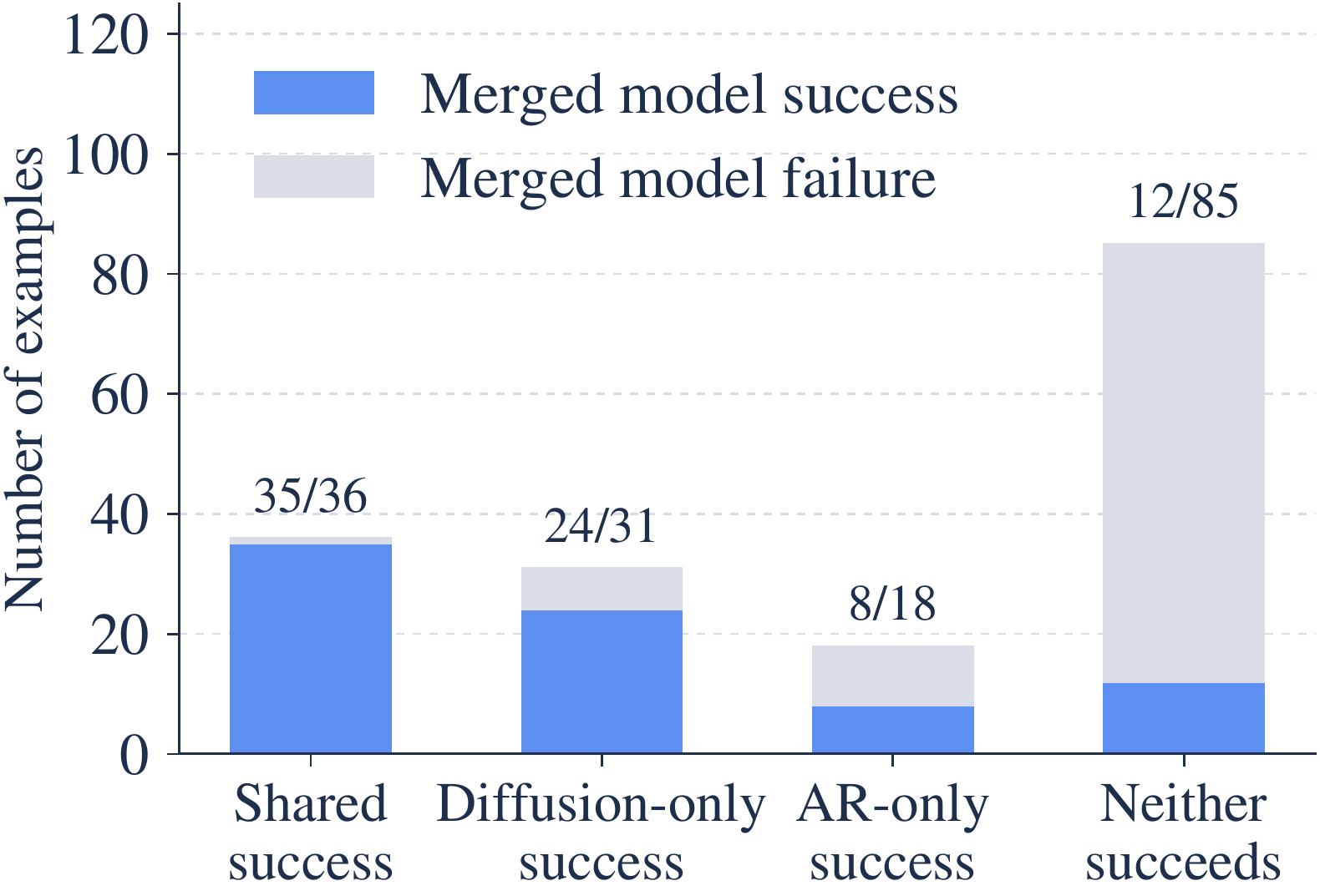}
        \caption{Interpolation preserves shared gains and endpoint-specific successes.}
        \label{fig:functional_disentanglement_pie}
    \end{subfigure}
    \caption{\textbf{Why AR and diffusion post-training can be composed.}
    \textbf{(a)} Performance over
    $\bm{D}_b+\alpha\bm{\tau}_A+\beta\bm{\tau}_D$.
    The interpolation path between the AR-transferred and diffusion-post-trained
    endpoints passes through high-performing regions and contains solutions that
    outperform both endpoints.
    \textbf{(b)} Among examples failed by $\bm{D}_b$, the two endpoints exhibit both
    shared and endpoint-specific successes. The best interpolated model preserves
    most shared successes while retaining gains unique to either endpoint,
    indicating limited functional interference under composition.}
    \label{fig:composition_analysis}
\end{figure}

As shown in ~\autoref{fig:composition_landscape}, both mathematical reasoning and coding exhibit high-performing regions along this interpolation path, and interior points along their interpolation path outperform either endpoint.
This indicates that interpolating the AR and diffusion post-training directions captures complementary gains beyond either endpoint.

\paragraph{Complementarity between AR and diffusion post-training.}
Why can an intermediate model outperform both endpoints?
To better understand these gains, we characterize the behavioral overlap and
differences between the two endpoints and the merged model.

Among examples failed by the base model $\bm{D}_b$ in mathematical reasoning, we partition problems into four
mutually exclusive subsets: those solved only by the AR-transfer model $\bm{D}_T$, only
by the post-trained model $\bm{D}_p$, by both, or by neither.
We then evaluate the best merged model on each subset.
As shown in \autoref{fig:functional_disentanglement_pie}, the two endpoints solve
overlapping but also distinct subsets of problems.
Importantly, the interpolated model preserves nearly all their shared successes
while retaining large portions of the successes unique to either endpoint.
Overall, we show that AR and diffusion post-training provide partially complementary
behavioral gains that can be jointly retained through interpolation, explaining
how an intermediate model can outperform both endpoints.
This observation is also consistent with prior findings that task-vector
composition can preserve distinct effects of different parameter-space
directions~\citep{ortiz2023task}.

\subsection{A2D: Cross-Paradigm Transfer and Composition}
\label{sec:a2d}

Motivated by the preceding observations, we propose \textbf{\methodname}
(\emph{AR-to-Diffusion Transfer and Composition}), a simple weight-space framework for recycling
AR post-training updates in diffusion models.
\methodname addresses two practical settings.

\paragraph{A2D Transfer.}
For many downstream capabilities, strong AR post-trained checkpoints are readily
available, while corresponding diffusion checkpoints may not yet exist.
In this setting, \methodname transfers the AR task vector directly to the diffusion base
to equip it with the target capability:
\begin{equation}
    \bm{D}_{\mathrm{Transfer}}(\alpha)
    =
    \bm{D}_b+\alpha\bm{\tau}_A,
    \qquad \alpha\geq0.
    \label{eq:a2d_transfer}
\end{equation}
This directly reuses existing AR post-training knowledge, avoiding a separate round of diffusion-specific post-training.

\paragraph{A2D Merge.}

For scenarios where both AR and diffusion post-trained models are available, \methodname interpolates
their post-training updates to exploit their complementary gains.
We restrict the composition to this one-dimensional path, which already
traverses a high-performing region in
\autoref{fig:composition_landscape}:
\begin{equation}
    \bm{D}_{\mathrm{Merge}}(\lambda)
    = \bm{D}_b + (1-\lambda)\boldsymbol{\bm{\tau}_D} + \lambda\bm{\tau}_A
    = \bm{D}_p + \lambda(\bm{\tau}_A-\bm{\tau}_D),
    \qquad \lambda\in[0,1].
\end{equation}
Both \methodname variants preserve the diffusion architecture and require no additional
training or inference-time hyperparameter tuning.

\section{A2D Across Models and Post-Training Settings}
\label{sec:experiments}
Beyond the controlled setting, we now evaluate \methodname on
independently released checkpoints, where the AR and diffusion branches may
differ substantially in their post-training data, objectives, and optimization
pipelines.
Our evaluation spans general instruction following, coding, mathematical
reasoning, and medical question answering, and covers both supervised and
reinforcement-learning-based post-training.
We organize the experiments by post-training type and evaluate both \methodname
regimes: \methodname-Transfer improves upon the diffusion base by directly applying an
AR post-training update, while \methodname-Merge further improves the diffusion
post-trained model by composing it with the corresponding AR update.

For each model family, the \methodname coefficient $\alpha$ and $\lambda$ is selected on a single anchor
benchmark and kept fixed across the evaluations of the remaining tasks.
We decode each diffusion model with its native inference settings and evaluate all code models with the same prompts and scoring. Parent diffusion checkpoints follow their original evaluation setup, while AR donors use the generation budget of the corresponding diffusion models.
Full evaluation, merging, and coefficient-selection details are provided in Appendix~\ref{app:experimental_details}.

\providecommand{\tbd}{{\color{red}\textbf{??}}}

\begin{table*}[t]
    \footnotesize
    \centering
    \setlength{\tabcolsep}{4pt}\renewcommand{\arraystretch}{1.1}

    \setlength{\fboxsep}{1.5pt}
    \caption{\textbf{Evaluation of \methodname across tasks and model families.}
    Headers denote \texttt{AR donor}$\rightarrow$\texttt{diffusion recipient}. \colorbox{blue!5}{\methodname\ (transfer)} applies the AR donor's task vector to the diffusion recipient, while \colorbox{blue!5}{\methodname\ (merge)} composes it with diffusion post-training.
    \colorbox{green!8}{Anchor tasks} for scaling coefficient search are highlighted.
    All results are in percentages. \looseness=-1}
    \label{tab:main}
    \vspace{-5pt}

    \begin{subtable}[t]{\linewidth}\centering
    \caption{Instruction following vectors.}
    \label{tab:main-ins}
    \resizebox{0.99\linewidth}{!}{%
    \begin{tabular}{@{}lccc!{\color{black!30}\vrule}ccc!{\color{black!30}\vrule}ccc@{}}
    \toprule
    & \multicolumn{3}{c!{\color{black!30}\vrule}}{Qwen2.5-7B $\rightarrow$ Dream-7B}
    & \multicolumn{3}{c!{\color{black!30}\vrule}}{Qwen2.5-Coder-7B $\rightarrow$ Dream-Coder}
    & \multicolumn{3}{c@{}}{Qwen2.5-Coder-7B $\rightarrow$ DiffuCoder} \\
    \cmidrule(lr){2-4}\cmidrule(lr){5-7}\cmidrule(l){8-10}
    & \colorbox{green!8}{IFEval} & MATH-500 & MBPP+
    & \colorbox{green!8}{HumanEval+} & MBPP+ & BCB-Full
    & \colorbox{green!8}{HumanEval+} & MBPP+ & BCB-Full \\
    \midrule
    \rowcolor{gray!5}
    AR donor
    & 75.3 & 36.6 & 69.0
    & 86.0 & 72.2 & 41.6
    & 86.0 & 72.2 & 41.8 \\
    \midrule
    Base
    & 49.7 & 20.8 & 59.3
    & 57.9 & 62.4 & 25.2
    & \textbf{61.6} & 52.9 & \textbf{33.9} \\
    \rowcolor{blue!5}
    \methodname\ (transfer)
    & \textbf{59.2} & \textbf{40.2} & \textbf{62.7}
    & \textbf{67.1} & \textbf{62.7} & \textbf{31.7}
    & 60.4 & \textbf{54.0} & 32.0 \\
    \midrule
    Instruct
    & 63.0 & 38.4 & 57.1
    & 77.4 & 68.5 & 36.2
    & 65.9 & 65.9 & 34.8 \\
    \rowcolor{blue!5}
    \methodname\ (merge)
    & \textbf{72.8} & \textbf{45.0} & \textbf{58.7}
    & \textbf{79.3} & \textbf{69.3} & \textbf{38.2}
    & \textbf{74.4} & \textbf{66.4} & \textbf{38.9} \\
    \bottomrule
    \end{tabular}}
    \end{subtable}
    \vspace{-3pt}

    \begin{subtable}[t]{\linewidth}\centering
    \caption{Reasoning, SFT and RL post-training vectors. DR abbreviates DreamReasoner.}
    \label{tab:main-others}
    \resizebox{0.99\linewidth}{!}{%
    \begin{tabular}{@{}lccc!{\color{black!30}\vrule}ccc!{\color{black!30}\vrule}ccc@{}}
    \toprule
    & \multicolumn{3}{c!{\color{black!30}\vrule}}{Qwen3-8B $\rightarrow$ DR-8B, thinking on}
    & \multicolumn{3}{c!{\color{black!30}\vrule}}{Gemma-4 $\rightarrow$ DiffusionGemma}
    & \multicolumn{3}{c@{}}{AceCoder-Rule $\rightarrow$ DiffuCoder, RL} \\
    \cmidrule(lr){2-4}\cmidrule(lr){5-7}\cmidrule(l){8-10}
    & \colorbox{green!8}{MATH-500} & AIME24 & AIME25
    & \colorbox{green!8}{VQA-RAD} & SLAKE & VQA-Med
    & HumanEval+ & \colorbox{green!8}{MBPP+} & BCB-Full \\
    \midrule
    \rowcolor{gray!5}
    AR donor
    & 81.2 & 72.1 & 60.4
    & 75.8 & 66.5 & 44.0
    & 85.4 & 70.6 & 43.3 \\
    \midrule
    Base
    & 68.0 & 5.8 & 12.1
    & 57.4 & 63.2 & 43.2
    & \textbf{65.9} & 65.9 & \textbf{34.8} \\
    \rowcolor{blue!5}
    \methodname\ (transfer)
    & \textbf{76.0} & \textbf{21.2} & \textbf{16.2}
    & \textbf{74.1} & \textbf{67.3} & \textbf{45.2}
    & 64.6 & \textbf{66.9} & 34.6 \\
    \midrule
    Post-trained
    & 93.6 & 71.7 & 56.7
    & 71.6 & 62.2 & 38.6
    & 63.4 & 66.9 & 40.5 \\
    \rowcolor{blue!5}
    \methodname\ (merge)
    & \textbf{94.8} & \textbf{72.9} & \textbf{67.1}
    & \textbf{74.5} & \textbf{63.8} & \textbf{38.8}
    & \textbf{70.1} & \textbf{69.0} & \textbf{42.4} \\
    \bottomrule
    \end{tabular}}
    \end{subtable}
 
    \vspace{8pt}

    \begin{subtable}[t]{\linewidth}
    \centering
    \caption{Ministral-3-3B $\rightarrow$ Nemotron-LD-3B, dual-mode AR-diffusion post-training vectors.
    The model supports both diffusion and AR decoding, and we report both results. The mean excludes the anchor task IFEval.}
    \label{tab:main-nemo}
    \resizebox{0.99\linewidth}{!}{%
    \begin{tabular}{@{}llcccccccccc|c@{}}
    \toprule
    & Mode & GSM8K & MATH-500 & HumanEval & MBPP
    & \colorbox{green!8}{IFEval} & MMLU & GPQA
    & AIME24 & AIME25 & LCB-CPP & Mean \\
    \midrule
    \rowcolor{gray!5}
    AR donor  & AR
    & 91.3 & 83.0 & 80.5 & 75.9 & 65.1
    & 67.9 & 40.9 & 30.0 & 13.3 & 27.1 & 56.7 \\
    \midrule
    
    Base & DLM
    & 59.1 & 40.0 & 64.6 & 70.9 & 23.2
    & 62.0 & \textbf{35.4} & 0.0 & 0.0 & 9.9 & 38.0 \\
    \rowcolor{blue!5}
    \methodname\ (transfer) & DLM
    & \textbf{88.1} & \textbf{52.6} & \textbf{72.6} & \textbf{74.3} & \textbf{60.6}
    & \textbf{62.8} & 33.3 & \textbf{3.3} & \textbf{3.3} & \textbf{14.1} & \textbf{44.9} \\
    \midrule
    Instruct & DLM
    & 89.2 & 76.8 & 73.2 & 73.0 & 68.0
    & 72.1 & 35.4 & 10.0 & \textbf{13.3} & 16.3 & 51.0 \\
    \rowcolor{blue!5}
    \methodname\ (merge) & DLM
    & \textbf{90.1} & \textbf{79.6} & \textbf{75.0} & \textbf{75.7}
    & \textbf{72.9} & \textbf{72.9} & \textbf{43.4}
    & \textbf{16.7} & \textbf{13.3} & \textbf{18.9} & \textbf{54.0} \\
    
    \midrule
    Instruct & AR
    & 87.6 & 76.0 & 73.8 & 69.8 & 68.8
    & 71.9 & \textbf{41.9} & \textbf{23.3} & \textbf{13.3} & 17.6 & 52.8 \\
    \rowcolor{blue!5}
    \methodname\ (merge) & AR
    & \textbf{89.3} & \textbf{81.0} & \textbf{79.9} & \textbf{75.1}
    & \textbf{72.8} & \textbf{72.5} & 38.4
    & 20.0 & \textbf{13.3} & \textbf{18.7} & \textbf{54.2} \\
    \bottomrule
    \end{tabular}}
    \end{subtable}
    \vspace{-10pt}

\end{table*}

\subsection{Instruction-Tuned Updates for General and Code Models}
\label{sec:exp_instruction}

We first evaluate whether instruction-tuning updates transfer from AR to diffusion across both
general-purpose and code-specialized model families.
For the general-purpose setting, we construct the AR instruction-tuning vector
by subtracting Qwen2.5-7B-Base from Qwen2.5-7B-Instruct~\citep{qwen25}, and
transfer this update to Dream-7B-Base~\citep{ye2025dream}.
We evaluate the resulting model on instruction following, mathematical
reasoning, and coding.

For the code-specialized setting, we similarly construct the instruction-tuning
vector from Qwen2.5-Coder-7B-Base and Qwen2.5-Coder-7B-Instruct
~\citep{hui2024qwen25coder}, and transfer it to two independently developed
diffusion code models, Dream-Coder~\citep{xie2025dreamcoder} and
DiffuCoder~\citep{gong2025diffucoder}.
In all three settings, the diffusion models originate from the corresponding AR
base family, while their instruction-tuned checkpoints are obtained through
separate diffusion-specific post-training pipelines.

As shown in \autoref{tab:main-ins}, \methodname is broadly effective across different types of instruction-following post-training, exhibiting two complementary regimes.
Direct transfer can recover substantial capability missing from the diffusion
base, with particularly large gains on Dream-7B for MATH-500 ($+19.4$), and on
Dream-Coder BCB-Full ($+6.5$).
When diffusion instruction tuning is already available, \methodname-Merge further provides
consistent gains, improving all $9/9$ benchmarks over the corresponding
post-trained baselines.
Moreover, scaling coefficients selected on a single anchor task generalize to the
remaining benchmarks, yielding average non-anchor gains of $4.8$ and $2.6$
points for transfer and merge, respectively.

\subsection{Specialized Post-Training: Reasoning, Medical VQA, and Code RL}
\label{sec:exp_reasoning}

We next test whether \methodname extends beyond general instruction tuning to
more specialized capabilities and later-stage post-training updates.
For mathematical reasoning, we use the post-training update from
Qwen3-8B-Base to its reasoning checkpoint~\citep{qwen3} and transfer it to
DreamReasoner-8B~\citep{wu2026dreamreasoner8bblocksizecurriculumlearning},
evaluating on MATH and competition-level math benchmarks.
For medical visual question answering (VQA), we use matched Gemma-4~\citep{gemmateam2026gemma4} and
DiffusionGemma medical-SFT updates, obtained by fine-tuning their respective
instruction-tuned checkpoints on VQA-RAD~\citep{van2026diffugemmavqa}, and evaluate on 3 medical VQA tasks. 
For code, we transfer and merge the RL updates from the instruction-tuned
checkpoints of AceCoder~\citep{zeng2025acecoder} and
DiffuCoder~\citep{gong2025diffucoder}.
Together, these settings cover both specialization from a base model and
continued SFT or RL applied after instruction tuning.

As shown in ~\autoref{tab:main-others}, \methodname remains effective across
reasoning, medical SFT, and code RL.
\methodname-Transfer yields substantial gains in the reasoning and medical settings,
including $+15.4$ on AIME24 and $+16.7$ on VQA-RAD.
\methodname-Merge is  consistently effective, improving all $9/9$ post-trained
baselines, with gains of $+10.4$ on AIME25 and $+6.7$ on HumanEval+.
These gains extend beyond the anchor tasks, averaging $4.0$ and $3.7$ points on
non-anchor benchmarks for transfer and merge, respectively.
Together, these results extend \methodname to specialized domain adaptation
and different stages of post-training, while showing robust performance
across model scales, as demonstrated by results for DiffusionGemma with 26B parameters. %

\subsection{A2D on Dual-Mode AR-Diffusion Models}
\label{sec:exp_nemotron}

Finally, we evaluate \methodname on Nemotron-Labs-Diffusion-3B (Nemotron-LD-3B)~\citep{fu2026nemotronlabsdiffusion}, a dual-mode model initialized from Ministral-3-3B~\citep{liu2026ministral3} that supports diffusion, AR, and self-speculative decoding. We evaluate \methodname-Transfer and \methodname-Merge across ten benchmarks covering math, coding, instruction following, and general reasoning under diffusion decoding, and additionally test the merged model under AR decoding. This setting also probes transfer with imperfect parameter correspondence, since Nemotron-LD contains diffusion-specific parameters with no AR counterpart. We keep these unmatched parameters from the diffusion recipient and apply the AR update only to aligned parameters. Full parameter-mapping details are provided in Appendix~\ref{app:merge}. 

As shown in \autoref{tab:main-nemo}, \methodname-Transfer improves the diffusion base on
$9/10$ benchmarks, with large gains on GSM8K
($+29.0$), MATH-500 ($+12.6$), and IFEval
($+37.4$), with a performance improvement of $6.9$ points averaged on the non-anchor tasks.
\methodname-Merge further improves or matches all $10/10$ benchmarks under diffusion
decoding, increasing the non-anchor mean from $51.0$ to $54.0$.
Notably, the same merged weights also remain effective under AR decoding, improving the
mean from $52.8$ to $54.2$, with notable gains on MATH-500 (+$5.0$), HumanEval (+$6.1$), and MBPP (+$5.3$).

Additionally, we also provide the \methodname-Transfer's performance on AR mode in~\autoref{tab:nemotron3b}, results with Nemotron-LD-8B in Appendix~\ref{app:nemontron8b}, which shows \methodname-Merge leads to notable gains when merging the reasoning post-training vector from Ministral-3-8B donor. Furthermore, we show that \methodname-merge generally improves the decoding efficiency for self-speculative decoding with Nemotron-LD in~\autoref{tab:nemotron-selfspec}. Together, these results show that \methodname demonstrates reliable performance improvement across a broad task
suite, tolerates imperfect parameter correspondence, and transfers its gains
across both diffusion and autoregressive decoding modes.

\section{Analysis and Ablations}
\label{sec:analysis}

We further analyze the behavior and scope of cross-paradigm transfer and
identify the sources of \methodname's gains.
Unless otherwise stated, experiments use Dream-7B on MATH-500, where we transfer the instruction-tuned vector from Qwen2.5-7B, and all scores
are reported in percentages.

\subsection{Understanding Cross-Paradigm Transfer}
\label{sec:analysis_transfer}

\begin{table*}[t]
    \centering
    \footnotesize
    
    \begin{minipage}[t]{0.265\textwidth}
    \vspace{-10pt}
    \centering
    \captionof{table}{
    A2D improves both denoising scheduling and token prediction.
    }
    \label{tab:lockstep-compact}
    \vspace{2pt}
    \setlength{\tabcolsep}{4.5pt}
    \renewcommand{\arraystretch}{1.05}
    \resizebox{\linewidth}{!}{%
    \begin{tabular}{@{}lcc@{}}
    \toprule
    & \multicolumn{2}{c}{Tokens from} \\
    \cmidrule(lr){2-3}
    Schedule from & Parent & A2D \\
    \midrule
    Parent & 38.4 & 42.8 \\
    A2D    & 42.2 & \textbf{45.0} \\
    \bottomrule
    \end{tabular}%
    }
    \end{minipage}
    \hfill
    \begin{minipage}[t]{0.31\textwidth}
    \vspace{-10pt}
    \centering
    \captionof{table}{
    The transfer is asymmetric: it transfers from AR to diffusion, but not in reverse.
    }
    \label{tab:reverse-main-small}
    
    \vspace{2pt}
    \setlength{\tabcolsep}{3.5pt}
    \renewcommand{\arraystretch}{1.05}
    
    \resizebox{\linewidth}{!}{%
    \begin{tabular}{@{}lcc@{}}
    \toprule
    & GSM8K & MATH-500 \\
    \midrule
    $D_b$
        & 66.5 & 20.8 \\
    \rowcolor{blue!5}
    $D_b$ + A2D transfer
        &\textbf{81.7} & \textbf{40.2} \\
    \midrule
    $A_b$
        & \textbf{62.5} & \textbf{17.4} \\
    \rowcolor{blue!5}
    $A_b$ + D2A transfer
        & 56.0 & 16.2 \\
    \bottomrule
    \end{tabular}%
    }
    \end{minipage}
    \hfill
    \begin{minipage}[t]{0.35\textwidth}
    \vspace{-10pt}
    \centering
    \captionof{table}{
    A2D benefits from both adding the AR vector and downscaling the diffusion vector.
    }
    \label{tab:gain-decomp-small}
    
    \vspace{2pt}
    \setlength{\tabcolsep}{4pt}
    \renewcommand{\arraystretch}{1.05}
    
    \resizebox{\linewidth}{!}{%
    \begin{tabular}{@{}lcc@{}}
    \toprule
    Method & Avg.\ Acc. & Avg.\ rank \\
    \midrule
    Diffusion parent $\ D_p$
        & 72.3 & 3.7 \\
    $+\tau_A$ only 
        & 75.9 & 2.4 \\
    $-\tau_D$ only 
        & 74.2 & 2.4 \\
    \rowcolor{blue!5}
    \methodname\-merge 
        & \textbf{76.8} & \textbf{1.4} \\
    \bottomrule
    \end{tabular}%
    }
    \end{minipage}
    \vspace{-10pt}
    \end{table*}

\paragraph{How does \methodname alter diffusion generation?}
We examine how \methodname affects the diffusion generation process to decompose its gains.

\textit{\methodname improves both denoising schedule and token prediction.}
As shown in \autoref{tab:lockstep-compact}, using \methodname-merged model token predictions with the denoising schedule of  Dream-7B-Instruct which we call here the \textit{diffusion parent}, improves
accuracy from $38.4$ to $42.8$, while using the merged model's denoising schedule with diffusion parent's
predictions reaches $42.2$.
Combining both yields the full \methodname performance of $45.0$, showing that
prediction and scheduling provide complementary gains.
These changes are nevertheless highly localized: when evaluated on the parent's
intermediate states, \methodname selects different positions at only $5.3\%$ of
steps and commits different tokens at only $3.4\%$ of positions compared to the diffusion parent.

\textit{\methodname does not systematically shift denoising toward causal
generation.}
Despite modifying the denoising schedule, \methodname does not induce a
systematic shift toward left-to-right decoding.
Following \citet{gong2025diffucoder}, we measure global and local AR-ness over
the response up to the first \texttt{EOS}. Global@$k$ measures whether the selected position is among the $k$ leftmost
masked tokens, Local@$k$ measures whether the current update continues a left-to-right run over the previous $k$ updates. Both equal $1$ under strict autoregressive decoding.
As shown in \autoref{fig:generation_behavior}(a) of the appendix, the merged model closely tracks the
diffusion parent on both metrics.

\textit{\methodname selectively revises uncertain predictions.}
The few changed tokens by \methodname are not spread evenly.
As shown in \autoref{fig:generation_behavior}(b), they concentrate at positions
where the diffusion parent has relatively low confidence.
Thus, \methodname largely preserves the parent's predictions while selectively
revising uncertain decisions. \looseness=-1

\paragraph{Directional asymmetry in cross-paradigm transfer.}
We test reversing the transfer direction and apply the diffusion task vector to the corresponding AR base.
As summarized in \autoref{tab:reverse-main-small} (full results in \autoref{tab:reverse-main}), while
applying AR task vector to diffusion base model substantially improves it, applying the diffusion task vector to the AR base model can degrade its performance, revealing a clear directional asymmetry. \looseness=-1

This asymmetry provides insight into what enables \methodname transfer.
The AR update $\bm{\tau}_A$ is learned around $\bm{A}_b$, from which $\bm{D}_b$ is subsequently
derived through diffusion conversion.
Thus, $\bm{D}_b$ inherits much of the parameter and representation context on which
$\bm{\tau}_A$ acts.
In contrast, $\bm{\tau}_D$ is learned around the already converted model $\bm{D}_b$ and
may depend on changes introduced during diffusion pretraining that are absent
from $\bm{A}_b$.
These results suggest that \methodname works in part because the diffusion recipient
shares the training lineage of the model on which the transferred AR update was
learned, whereas this compatibility is absent in the reverse direction.

\paragraph{Scope of cross-paradigm transfer.}
Our main experiments consider task vectors induced by downstream fine-tuning,
for which \methodname broadly enables cross-paradigm transfer.
We additionally test vectors induced by continued pretraining, respectively from
Qwen2.5-7B to Qwen2.5-Math-7B and Qwen2.5-Coder-7B, and find that neither transfers
effectively to the corresponding diffusion model.
These vectors have $70$--$82\times$ larger norms than the post-training vectors
used in our main experiments, reflecting substantially broader parameter
changes that may reorganize the representations on which the updates rely.
These results delineate \methodname's scope to relatively local post-training updates,
rather than pretraining-scale model transformations.

\subsection{Ablation Studies}
\label{sec:ablations}

\paragraph{Decomposition of the \methodname gain.}
\methodname-Merge can be written as
$\bm{D}_p+\lambda(\bm{\tau}_A-\bm{\tau}_D)$, combining two effects: adding the AR update and
partially removing the diffusion update.
We isolate these effects by retaining only $+\bm{\tau}_A$ or $-\bm{\tau}_D$ contribution with a corresponding scaling coefficient.
As shown in \autoref{tab:gain-decomp-small} (full results in
Table~\ref{tab:gain-decomp}), both components individually improve over the
diffusion parent, but neither matches \methodname-Merge.
Thus, the full gain benefits from jointly balancing the AR and diffusion
post-training updates. \looseness=-1

\paragraph{Module-wise contributions to \methodname.}
We ablate the contribution of attention and MLP modules from $\bm{\tau}_A$.
As shown in \autoref{tab:module_ablation}, while \methodname-Transfer depends mostly on updates in MLP modules, \methodname-Merge requires both MLP and attention modules to achieve optimal merging performance. \looseness=-1

\section{Related Work}
\label{sec:related_work}

\vspace{-4pt}
\paragraph{Diffusion large language models} 
\looseness=-1

Extending diffusion-based generative modeling to discrete sequences has been widely studied~\citep{austin2021structured,hoogeboom2021argmax,han2022ssd,gong2022diffuseq}, with masked diffusion emerging as a particularly effective formulation for language modeling~\citep{he2023diffusionbert,sahoo2024simple,shi2024simplified}. Subsequent scaling efforts established diffusion language models (dLLMs) as competitive alternatives to autoregressive (AR) models~\citep{nie2025scaling,nie2025large,ye2025dream,song2025seed,gong2025diffucoder,wu2026dreamreasoner8bblocksizecurriculumlearning}. Rather than training from scratch, many recent dLLMs build on pretrained AR checkpoints and introduce diffusion objectives through continued training or adaptation~\citep{gong2025scaling,ye2025dream,xie2025dreamcoder,wu2026fast,cheng2026sdar,fu2026nemotronlabsdiffusion,diffugemma2026technicalreport}, reusing pretrained linguistic knowledge while substantially reducing the cost of learning diffusion-based generation. Post-training dLLMs, however, poses distinct optimization challenges as learning must span partially corrupted states across varying corruption levels~\citep{parashar2026learnability, deng2026beyond}, with additional challenges in likelihood estimation for RL~\citep{gong2025diffucoder, tang2026gdsd}.

Recent work further reveals shared structure between the two paradigms: AR models can guide diffusion adaptation~\citep{peng2026don,su2026data}, while left-to-right biases can remain beneficial even with flexible generation orders~\citep{xue2025any,ni2026flexibility}. Together, these results suggest that AR and diffusion models share reusable knowledge while retaining distinct inductive biases. We study this connection in \textit{weight space}, asking whether AR post-training updates remain effective after diffusion conversion and can be composed with diffusion updates.

\vspace{-4pt}
\looseness=-1

\paragraph{Model merging.}
Model merging aims to combine the capabilities of independently trained models
directly in weight space, avoiding additional training \citep{yang2026model}.
Early work showed that models fine-tuned from a common initialization often
remain connected in weight space for parameter averaging to improve
performance, motivating both simple averaging \citep{wortsman2022modelsoup} and importance-weighted variants \citep{matena2022fisher, tam2023mats}.
Task arithmetic formalized fine-tuning updates as task vectors that
can be composed in weight space to manipulate model behavior
\citep{ilharco2022editing}. 
Its effectiveness has been linked to functional disentanglement among task
vectors~\citep{ortiz2023task}, with subsequent work mitigating interference
through conflict resolution~\citep{yadav2023ties}, sparsification
~\citep{yu2024dare,wang2024localizing}, and layer-wise rescaling
~\citep{wang2025lines,yang2024adamerging}.

Prior work on LLMs primarily focused on model merging within the AR paradigm~\citep{ren2026enough, zhu2026sftconflicts}, whereas we study \textit{cross-paradigm transfer} from AR to diffusion. Closest to our setting, \citet{lin2026ar} transfers AR preference-alignment updates as an alternative to diffusion-based alignment, while we study a broader post-training setting where AR knowledge is used to further improve already post-trained diffusion models across diverse objectives, capabilities, and model families.

\section{Conclusion}
\label{sec:conclusion}

We show that AR post-training knowledge exhibits \textit{cross-paradigm transfer}: it remains effective after diffusion conversion
and can be reused directly in diffusion models through weight-space transfer.
Building on this cross-paradigm transferability, \methodname transfers AR post-training
updates to diffusion models and, when diffusion post-training is available,
combines the two to exploit complementary gains. Across diverse model families,
tasks, and post-training settings, \methodname consistently improves diffusion models
without additional training or inference-time computation.

Practically, \methodname provides a simple training-free framework for reusing existing AR post-trained
checkpoints as resources for diffusion models, supporting both capability
transfer to diffusion base models and further improvement of already
post-trained diffusion models through composition.

\subsubsection*{Acknowledgments}
\label{sec:acknowledgements}
Yiming Qin was supported by the Swiss National Science Foundation (SNSF grant 10001445). 
The authors thank Nikolaos Dimitriadis and Guillermo Ortiz-Jiménez for their constructive feedback on the project, and Zhengqing Wu and Keyue Jiang for insightful discussions.

\bibliography{tools/ref}
\bibliographystyle{iclr2027_conference}

\newpage

\addcontentsline{toc}{section}{Appendix} %
\part{Appendix} %
\parttoc %

\appendix
\clearpage
\section{Experimental Details}\label{app:experimental_details}

This section describes how each model family is evaluated, how the merges are built, and how the coefficients are chosen.

\subsection{Models and the evaluation pipeline}
\label{app:models_evals}

Every diffusion model in this work is initialized from an AR model, as illustrated in 
Sec.~\ref{sec:motivation}. Each family has an AR base and its post-trained model, and a diffusion base
$\bm{D}_b$ and its post-trained model $\bm{D}_p$. We call the AR post-trained model the \emph{AR donor}, because its
task vector $\bm{\tau}_A$ is the one we transfer, and we call $\bm{D}_p$ the \emph{diffusion parent}, as in \autoref{sec:analysis}, because it is the model A2D is built on and compared against. We take these relationships from each release's paper and model card. Diffusion models and their A2D variants use the diffusion model's official evaluation hyperparameters, and we reproduce each parent's published numbers before evaluating any merge. Each family below gives its evaluation protocol and, specifically, every point at which we
depart from the official description of the parent model.

\paragraph{Dream-7B~\citep{ye2025dream} / Qwen2.5-7B~\citep{qwen25}.}Dream-7B adapts Qwen2.5-7B with a diffusion objective, and Dream-Instruct adds SFT on instruction pairs. IFEval,
GSM8K, MATH-500 and HumanEval run in the Dream fork of \texttt{lm-evaluation-harness} at its released decoding settings
and per-task budgets, with the chat template and one denoising step per generated token. IFEval is the mean of the four
lm-eval metrics, and GSM8K and MATH-500 are 0-shot. MATH-500 replaces the release's full MATH task at its $512$-token
budget. HumanEval+ scores the generations of the
fork's \texttt{humaneval\_instruct} task with the EvalPlus~\citep{liu2023evalplus} tests, and MBPP+ uses the Dream-Coder EvalPlus harness at Dream's own $1{,}024$-token budget.
IFBench is included to provide a more challenging instruction following benchmark~\citep{pyatkin2026ifbench}, generated with the same decoding settings at $1{,}280$
tokens and scored with the official IFBench code at prompt-level loose accuracy.

\textit{GSM8K prompt.}
On GSM8K we add the system prompt \emph{``Solve the math problem with brief, direct reasoning. Do NOT restate the problem, do NOT use bullet points, and use as few steps as possible. On the final line, write the answer as `\#\#\#\# $<$number$>$' and nothing else.''} Without the prompt the model often runs out of the $256$-token budget before it reaches the final-answer format. The prompt helps the parent as well, and we use it for every Dream-7B GSM8K evaluation. 
The AR donor is evaluated with the same prompt for fairness.

\paragraph{Dream-Coder~\citep{xie2025dreamcoder} / Qwen2.5-Coder~\citep{hui2024qwen25coder}.}
Dream-Coder-7B is built from Qwen2.5-Coder-7B, and Dream-Coder-Instruct applies SFT followed by GRPO, so its task
vector carries both. HumanEval, MBPP and  BigCodeBench(BCB) use the EvalPlus~\citep{liu2023evalplus} fork released with Dream-Coder at its documented settings. We report HumanEval+ and MBPP+, and BCB runs on the Instruct split (full subset), following the default setting of the official release for DreamCoder.

\paragraph{DiffuCoder~\citep{gong2025diffucoder} / Qwen2.5-Coder~\citep{hui2024qwen25coder} / AceCoder-Rule~\citep{zeng2025acecoder}.}
DiffuCoder-7B continues Qwen2.5-Coder-7B with diffusion pre-training on code, DiffuCoder-Instruct adds SFT, and DiffuCoder-cpGRPO adds coupled-GRPO. This gives two task vectors on one architecture: an SFT vector from Base to Instruct, paired with Qwen2.5-Coder-7B to Qwen2.5-Coder-7B-Instruct, and an RL vector from Instruct to cpGRPO, paired with Qwen2.5-Coder-7B-Instruct to AceCoder-Rule.
We use the Dream-style EvalPlus harness with DiffuCoder's own inference hyperparameters, such as temperature and denoising budget. DiffuCoder reports the best of $T\in\{0.2,0.3,0.4\}$ for the RL checkpoint, whereas we compare every variant at its default $T{=}0.2$. BCB runs on the Complete split, as reported in the main results of DiffuCoder, and we use Dream-Coder's harness, with $512$ steps and top-$p$ $0.95$, as suggested in DiffuCoder.

\paragraph{DreamReasoner~\citep{wu2026dreamreasoner8bblocksizecurriculumlearning} / Qwen3~\citep{qwen3}.}
DreamReasoner-8B is initialized from Qwen3-8B-Base. We use the
release's block-diffusion decoder at its documented settings, with thinking enabled unless stated otherwise. AIME24/25
follow the paper, which averages eight runs at a $24{,}576$-token budget because the output varies between runs even at
$T{=}0$, and we report that mean as avg@$8$. MATH-500 uses a single greedy sample.

\textit{Base evaluation.}
The DreamReasoner chat template wraps the answer in \texttt{<think>} markers that the base model has never seen, and it repeats them until the budget runs out. We therefore give the base model the plain problem with the same \texttt{\textbackslash boxed\{\}} instruction. The transfer model reads the template normally, and we evaluate it in both modes.

\paragraph{Nemotron-Labs-Diffusion~\citep{fu2026nemotronlabsdiffusion} / Ministral~3~\citep{liu2026ministral3}.}
The 3B and 8B Nemotron-LD models are initialized from Ministral~3 Base. The four checkpoints are Nemotron-LD Base/Instruct
and Ministral~3 Base-2512/Instruct-2512, and merges into the Instruct models retain the recipient's tokenizer and chat template. Base-transfer models use the interface described below.
We use NeMo-Skills for all evaluations. For the Ministral donor checkpoints, we sample one response per problem with temperature $0.7$, top-$p$ $0.95$, and a maximum of $8{,}192$ output tokens. The donor evaluations use native Mistral chat formatting and benchmark-provided system instructions.

\textit{Base evaluation.}
Ideally the diffusion Base would use the same instruction-following interface as the post-trained models.
Nemotron-LD-8B-Base does not have that interface yet. Its embedding rows for the chat markers
\texttt{<|im\_start|>} and \texttt{<|im\_end|>} are exactly zero, because those ids sit on Ministral's
\texttt{[IMG]} and \texttt{<pad>} slots and are trained only at the Instruct stage. With no signal on the turn
markers the Base cannot tell where the user turn ends, so it writes the role labels out as plain text and repeats
them. It still answers math, but drops sharply on code, GPQA, IFEval and LiveCodeBench
(Table~\ref{tab:nemotron8b}). That score reflects a missing interface rather than the Base's ability, and using it
as a baseline would inflate the gains we report. We therefore use the raw prompt, which the Base can read, in the
main results and give both settings in the appendix. We evaluate it on the raw problem text at its native
positional setting, with EOS set to ID~2 (\texttt{</s>}) as in its generation configuration.

\textit{Transfer evaluation.}
The transfer model cannot use the Nemotron chat template either, for the same reason. Those two rows are among the
mismatched rows frozen at Base (App.~\ref{app:merge}), and the AR vector does not rewrite them. What the AR vector
does bring is the donor's own interface. \texttt{[INST]}, \texttt{[/INST]} and \texttt{</s>} are IDs 3, 4 and 2,
and these ids carry the same strings in both tokenizers, so they do receive the AR update. We therefore prompt the
transfer rows in the donor's format, \texttt{<s>[INST]\{problem\}[/INST]}, with no system prompt and no thinking
marker. The difference is large. On the same 20 IFEval prompts at $\lambda{=}1.0$, the Nemotron template leaves 12 of
the 20 generations looping and scores $17.9$, while \texttt{[INST]} leaves 1 and scores $60.8$. The 3B transfer
uses the same donor format, although its Base chat-marker embeddings are not zero.

\textit{Reasoner model for Nemotron-LD-8B}
For the 8B model we also consider the Ministral-3-8B-Reasoning donor, because the 8B
diffusion model reasons on its own even when thinking is disabled. On LiveCodeBench (LCB) it emits
\texttt{</think>} by itself in $15.0\%$ of non-thinking generations, against $0\%$ for the 3B model. A
further $60.8\%$ of its generations carry a \texttt{</think>} that the server inserts at generated
token $6{,}000$, a length artefact rather than a behaviour. The thinking-off comparison with
an Instruct donor is therefore not fully matched at 8B.
Unless otherwise stated, evaluations use $8{,}192$ output tokens
and the Instruct positional settings. 

\paragraph{DiffusionGemma~\citep{diffugemma2026technicalreport} / Gemma~4~\citep{gemmateam2026gemma4}.}
DiffusionGemma is initialized from Gemma~4-26B-A4B-Instruct and trained by supervised diffusion adaptation, sampler
distillation, and reinforcement learning. It has no public base checkpoint, so for this pair we study domain-specific
fine-tuning instead of the Base/Instruct setup, using the Gemma~4 and DiffusionGemma fine-tunes released by
\citet{van2026diffugemmavqa} for SLAKE~\citep{liu2021slake}, VQA-RAD~\citep{lau2018vqarad}, and
VQA-Med-2019~\citep{benabacha2019vqamed}, with the two instruction-tuned checkpoints as the pre-task references.

We use the Transformers implementation of the native block-diffusion decoder with block length $256$, at most $48$
denoising steps per block, entropy bound $0.1$, and temperature annealed from $0.8$ to $0.4$, stopping early once the
predictions stop changing between steps. Thinking is disabled. All three datasets are evaluated 0-shot with the image
followed by the question and a $256$-token budget, on all $1{,}061$ English SLAKE test questions, all $451$ public
VQA-RAD test questions, and all $500$ VQA-Med-2019 test questions. We report semantic answer accuracy judged by
Qwen3.8-27B~\citep{qwen2026qwen38} with greedy decoding and thinking disabled. Unlike \citet{van2026diffugemmavqa}, which evaluated 350 random questions per dataset with Claude Sonnet-4.6, we use all test questions and an open judge, so their results serve only as reference.

\subsection{Details for model merging}
\label{app:merge}

\paragraph{Qwen-derived models.}
For Dream-7B, Dream-Coder, DiffuCoder, and DreamReasoner the diffusion model keeps its AR parent's architecture,
parameter names, and tensor shapes, so the merge is a direct element-wise combination of all parameters, with no
remapping, freezing, or masking.

Tokens that exist only in the diffusion tokenizer need no special treatment. The mask token and the other ids beyond the
AR vocabulary never occur in the AR model's training data, so $\bm{\tau}_A$ is exactly zero on their input-embedding rows for the
Qwen2.5-based vectors and small for Qwen3. We merge them with the same formula as every other row.

\paragraph{Nemotron / Ministral.}
\begin{table}[t]
\centering\footnotesize
\caption{\textbf{Special-token mismatches across the four 3B checkpoints.} All listed input-embedding rows are masked during task-vector composition.}
\label{tab:tokenizer-mismatches-3b}
\resizebox{\linewidth}{!}{%
\begin{tabular}{@{}rllll@{}}
\toprule
ID & Ministral Base & Ministral Instruct & Nemotron Base & Nemotron Instruct \\
\midrule
10 & \texttt{[IMG]} & \texttt{[IMG]} & \texttt{<|im\_start|>} & \texttt{<|im\_start|>} \\
11 & \texttt{<pad>} & \texttt{<pad>} & \texttt{<|im\_end|>} & \texttt{<|im\_end|>} \\
12 & \texttt{[IMG\_BREAK]} & \texttt{[IMG\_BREAK]} & \texttt{<think>} & \texttt{<think>} \\
13 & \texttt{[IMG\_END]} & \texttt{[IMG\_END]} & \texttt{</think>} & \texttt{</think>} \\
14 & \texttt{[PREFIX]} & \texttt{[PREFIX]} & \texttt{<tool\_call>} & \texttt{<tool\_call>} \\
15 & \texttt{[MIDDLE]} & \texttt{[MIDDLE]} & \texttt{</tool\_call>} & \texttt{</tool\_call>} \\
16 & \texttt{[SUFFIX]} & \texttt{[SUFFIX]} & \texttt{<tool\_response>} & \texttt{<tool\_response>} \\
17 & \texttt{[SYSTEM\_PROMPT]} & \texttt{[SYSTEM\_PROMPT]} & \texttt{</tool\_response>} & \texttt{</tool\_response>} \\
18 & \texttt{[/SYSTEM\_PROMPT]} & \texttt{[/SYSTEM\_PROMPT]} & \texttt{<SPECIAL\_18>} & \texttt{<SPECIAL\_18>} \\
19 & \texttt{[TOOL\_CONTENT]} & \texttt{[TOOL\_CONTENT]} & \texttt{<SPECIAL\_19>} & \texttt{<SPECIAL\_19>} \\
24 & \texttt{[AUDIO]} & \texttt{[AUDIO]} & \texttt{<SPECIAL\_24>} & \texttt{<SPECIAL\_24>} \\
25 & \texttt{[BEGIN\_AUDIO]} & \texttt{[BEGIN\_AUDIO]} & \texttt{<SPECIAL\_25>} & \texttt{<SPECIAL\_25>} \\
32 & \texttt{[ARGS]} & \texttt{[ARGS]} & \texttt{<SPECIAL\_32>} & \texttt{<SPECIAL\_32>} \\
33 & \texttt{[CALL\_ID]} & \texttt{[CALL\_ID]} & \texttt{<SPECIAL\_33>} & \texttt{<SPECIAL\_33>} \\
34 & \texttt{[THINK]} & \texttt{<SPECIAL\_34>} & \texttt{<SPECIAL\_34>} & \texttt{<SPECIAL\_34>} \\
35 & \texttt{[/THINK]} & \texttt{<SPECIAL\_35>} & \texttt{<SPECIAL\_35>} & \texttt{<SPECIAL\_35>} \\
\bottomrule
\end{tabular}}
\end{table}

Nemotron needs an explicit mapping from Ministral's language-model parameters to the diffusion backbone. The vocabulary
sizes agree, but 14 token IDs denote different tokens in the two families, and IDs 34--35 additionally differ between
Ministral Base and Instruct. Some of these are generation controls: ID 11 is padding in Ministral but end-of-turn in
Nemotron, and IDs 12--13 are image delimiters in Ministral but thinking delimiters in Nemotron
(Table~\ref{tab:tokenizer-mismatches-3b}). 
For protected Instruct merges, we keep these 16 input-embedding rows
at their values in the Nemotron Instruct recipient, excluding them
from both the AR-vector addition and the diffusion-vector subtraction.
For corrected Base-transfer experiments, we additionally freeze the
diffusion mask-token embedding (ID~100), keeping all 17 protected rows
at their Nemotron Base values.

The diffusion output head is likewise kept at its value in the recipient checkpoint (Base for transfer into Base, Instruct
for merges into Instruct), so donor updates enter through the mapped backbone and the compatible input embeddings without
touching the diffusion readout. 
Instruct merges retain the Nemotron Instruct evaluation interface,
whereas Base-transfer models use the donor-format \texttt{[INST]}
interface. Prompt formats, positional settings, and generation budgets
for these models and their Base baselines are specified in
App.~\ref{app:models_evals}.

\paragraph{DiffusionGemma / Gemma~4.}
We map the AR updates to the corresponding DiffusionGemma language-model projections. The diffusion encoder and decoder
share their base weights, so each AR update is applied once to the shared parameter and affects both branches. The
diffusion fine-tuning adapters, which hold separate encoder and decoder updates, stay unmerged and are scaled according to
the experimental formula. Diffusion-specific adapter updates such as the self-conditioning projections have no AR
counterpart.
The vision encoder,
embeddings, normalization parameters, and output head are unchanged, and all merged models keep the diffusion recipient's
tokenizer and chat template.

\subsection{Experimental details for \autoref{sec:motivation}}
\label{app:motivation_details}

\paragraph{Training.}
For mathematical reasoning we fine-tune Qwen2.5-7B and Dream-Base-7B on 100K NuminaMath-CoT
\citep{numina_math_datasets} examples; for coding we fine-tune Qwen2.5-Coder-7B and Dream-Coder-Base-7B on 100K
EpiCoder-func \citep{wang2025epicoder} examples. Training uses AdamW ($\beta_1=0.9$, $\beta_2=0.95$, weight decay $0.01$),
learning rates of $5\times10^{-6}$ for the AR models and $2\times10^{-6}$ for Dream, cosine decay with 10\% warmup, batch
size 256, and one epoch (390 updates), with sequences right-truncated to 2,048 tokens. Prompts use each model's single-user
chat template and the loss covers only the response tokens. AR models use causal next-token cross-entropy. Dream uses the
bidirectional masked-denoising loss with CART time reweighting and a random per-batch sequence cutoff.

\paragraph{Evaluation.}
We evaluate 0-shot on GSM8K, MATH-500, HumanEval+, and MBPP+, reporting answer accuracy for math and pass@1 for code,
including the extended EvalPlus tests. To isolate the effects of fine-tuning, we evaluate Dream using the same chat template as during training, without additional system instructions or assistant prefilling.
Specifically, we apply Dream's native chat template to a single user message,
retaining its default system message (``You are a helpful assistant.'')
and setting \texttt{add\_generation\_prompt=True}.
We add no custom system message or assistant continuation beyond
the template's generation prefix.
For MATH-500, the user message contains the original problem.
For GSM8K, it contains the question and the additional instruction,
formatted as \texttt{Q: \{question\}}, followed by the instruction for concise reasoning as in Section \ref{app:models_evals}, and an \texttt{A:} suffix.
Because fine-tuning lengthens responses, the budget is 512 tokens for GSM8K and coding and 1,024 for
MATH-500. Dream uses diffusion-step budgets equal to the generation-token
budgets,, entropy-based unmasking, temperature $0.1$, and top-$p$
$0.9$. 
For GSM8K, we extract a numeric answer from the final
\texttt{\textbackslash boxed\{\}} expression when parseable;
otherwise, we use the harness's flexible numeric extraction.
For MATH-500, we first use the Minerva boxed-answer and
final-answer-pattern extraction. If these fail, we apply a
Math-Verify-based prose extraction fallback.
Predictions for MATH-500 are then evaluated using the Minerva normalization
and symbolic-equivalence routine; Math-Verify is used for
fallback extraction. For coding, the user message contains the benchmark task prompt,
without additional coding instructions or an assistant code prefill.
We apply the same frozen scoring implementation to all compared models. 
We construct merged checkpoints using sequential CPU FP32 arithmetic, followed by a single cast to BF16, retaining the Dream base model’s configuration and tokenizer.
Math sweeps use first 200 examples in dataset order per benchmark unless stated otherwise, and coding evaluations use all 164 HumanEval
and 378 MBPP problems.

\subsection{GPU and training time}
All evaluations run on a single H100, H200, or B300 node. Because these pools do not give identical numbers, every run in a
hyperparameter sweep uses the same hardware, and so do directly compared variants (base and transfer, post-trained and
merged) on each benchmark. The four fine-tuning runs of \autoref{sec:motivation} take 30.8 GPU-hours in total, 6.1--9.2
GPU-hours each on four H100/H200 GPUs.

\subsection{Hyperparameter search}

For each family and construction, we fix an anchor benchmark
(Table~\ref{tab:sweep-anchors}).
For example, for
instruction following the anchor is IFEval. The coefficient is searched on the anchor only and then kept fixed on every
other benchmark, so the main table reports the anchor alongside tasks that played no role in the selection. For DiffuCoder, we use HumanEval+ as the anchor for instruction tuning, consistent with the other coding models, and MBPP+ for RL because we do not observe a gain from the instruction-tuned model to the RL model on HumanEval+ in our setup.

\begin{table}[ht]
\centering\footnotesize
\setlength{\tabcolsep}{5pt}
\renewcommand{\arraystretch}{1.15}
\caption{\textbf{Coefficient selection for each transfer family.}
Each coefficient is selected on its anchor benchmark and kept fixed
on all other benchmarks. For the Nemotron Reasoning-donor row, the
anchors are listed in transfer / merge order: IFEval without thinking
and MATH-500 with thinking, respectively.
Full sweeps are reported in Tables~\ref{tab:sweep-transfer} and \ref{tab:sweep-anchors}.}
\label{tab:hparam-chosen}
\resizebox{1.0\linewidth}{!}{%
\begin{tabular}{@{}l l l c c@{}}
\toprule
AR $\rightarrow$ Diffusion & vector type & anchor & transfer $\alpha$ & merge $\lambda$ \\
\midrule
Qwen2.5-7B $\rightarrow$ Dream-7B & instruct (SFT) & IFEval & 0.9 & 0.6 \\
Qwen2.5-Coder $\rightarrow$ Dream-Coder & code (SFT) & HumanEval+ & 0.6 & 0.2 \\
Qwen2.5-Coder $\rightarrow$ DiffuCoder & code (SFT) & HumanEval+ & 0.1 & 0.2 \\
AceCoder-Rule $\rightarrow$ DiffuCoder & code (RL) & MBPP+ & 0.2 & 0.5 \\
Qwen3-8B $\rightarrow$ DreamReasoner & reasoning & MATH-500 & 0.5 & 0.2 \\
Ministral-3-3B $\rightarrow$ Nemotron-LD-3B & instruct (SFT) & IFEval & 0.9 & 0.3 \\
Ministral-3-8B $\rightarrow$ Nemotron-LD-8B & instruct (SFT) & IFEval & 1.4 & 0.4 \\
Ministral-3-8B-Reasoning $\rightarrow$ Nemotron-LD-8B
& reasoning
& IFEval / MATH-500 
& 0.8 & 0.2 \\
Gemma~4 $\rightarrow$ DiffusionGemma & SLAKE (SFT) & SLAKE & 1.0 & 0.1 \\
Gemma~4 $\rightarrow$ DiffusionGemma & VQA-RAD (SFT) & VQA-RAD & 0.7 & 0.3 \\
\bottomrule
\end{tabular}}
\end{table}

Full sweeps for the main results and the key ablations are in Table~\ref{tab:sweep-anchors}, ~\ref{tab:sweep-addition}, ~\ref{tab:sweep-rescale} and ~\ref{tab:sweep-transfer}.

\providecommand{\sel}[1]{\cellcolor{blue!8}#1}
\providecommand{\tbd}{{\color{red}\textbf{??}}}

\begin{table*}[ht]
\footnotesize\centering
\setlength{\tabcolsep}{4pt}\renewcommand{\arraystretch}{1.1}
\caption{\textbf{Merge}, $D_b+(1-\lambda)\tau_D+\lambda\tau_A$, swept on each family's anchor benchmark. For DiffusionGemma, $D_b$ denotes the instruction-tuned reference checkpoint. \colorbox{blue!8}{Shaded} = selected $\lambda$, \textbf{bold} = best score. All results are in percentages.
}
\label{tab:sweep-anchors}
\resizebox{0.98\linewidth}{!}{%
\begin{tabular}{@{}l l cccccccccc@{}}
\toprule
AR $\rightarrow$ Diffusion
& Anchor
& \multicolumn{10}{c@{}}{$\lambda$} \\
\cmidrule(l){3-12}
& & $0$ (parent) & $0.1$ & $0.2$ & $0.3$ & $0.4$
& $0.5$ & $0.6$ & $0.7$ & $0.8$ & $1.0$ \\
\midrule

Qwen2.5-7B $\rightarrow$ Dream-7B
& IFEval
& 63.0 & 65.0 & 67.6 & --- & 69.8 & 70.3 & \sel{\textbf{72.8}} & 72.6 & 71.5 & 56.7 \\

Qwen2.5-Coder $\rightarrow$ Dream-Coder
& HumanEval+
& 77.4 & 76.2 & \sel{\textbf{79.3}} & 76.2 & 77.4
& --- & --- & --- & --- & --- \\

Qwen2.5-Coder $\rightarrow$ DiffuCoder
& HumanEval+
& 65.9 & 70.7 & \sel{\textbf{74.4}} & 72.6 & 71.3
& --- & --- & --- & --- & --- \\

AceCoder-Rule $\rightarrow$ DiffuCoder (RL)
& MBPP+
& 66.9 & --- & 66.4 & \textbf{69.0} & 68.3
& \sel{\textbf{69.0}} & 66.9 & --- & --- & --- \\

Qwen3-8B $\rightarrow$ DreamReasoner
& MATH-500
& 93.6 & 93.2 & \sel{\textbf{94.8}} & 94.0 & 94.2
& --- & --- & --- & --- & --- \\

Ministral-3-3B-Instruct $\rightarrow$ Nemotron-LD-3B
& IFEval
& 68.0 & --- & 71.5 & \sel{\textbf{72.9}} & 71.9 & 72.5 & 69.1 & --- & 51.2 & --- \\

\addlinespace[1pt]
Ministral-3-8B-Instruct $\rightarrow$ Nemotron-LD-8B
& IFEval
& 69.8 & --- & 74.4 & 74.1 & \sel{\textbf{78.8}} & 76.2 & --- & --- & --- & --- \\

Ministral-3-8B-Reasoning $\rightarrow$ Nemotron-LD-8B
& MATH-500
& 87.2 & 90.8 & \sel{\textbf{92.6}} & 90.4 & 90.2 & --- & 39.4 & --- & 11.4 & --- \\

Gemma~4 $\rightarrow$ DiffusionGemma
& SLAKE
& 83.1 & \sel{\textbf{83.3}} & 82.9 & --- & 82.3
& --- & 81.8 & --- & 81.2 & 77.1 \\

Gemma~4 $\rightarrow$ DiffusionGemma
& VQA-RAD
& 71.6 & 72.7 & 73.8 & \sel{\textbf{74.5}} & 73.6
& --- & 68.7 & --- & 71.8 & 70.5 \\

\bottomrule
\end{tabular}}
\end{table*}

\begin{table*}[ht]
    \footnotesize\centering
    \setlength{\tabcolsep}{5pt}\renewcommand{\arraystretch}{1.1}
    \caption{\textbf{Transfer}, $D_b+\lambda\tau_A$: the AR vector on the diffusion base. For DiffusionGemma, $D_b$ denotes the instruction-tuned reference checkpoint. Shading and bold as above; $\lambda{=}0$ is the base. All results are in percentages.
    }
    \label{tab:sweep-transfer}
    \resizebox{0.98\linewidth}{!}{%
    \begin{tabular}{@{}l l ccccccccccccccc@{}}
    \toprule
    AR $\rightarrow$ Diffusion & Anchor & \multicolumn{15}{c@{}}{$\alpha$} \\
    \cmidrule(l){3-17}
    & & $0$ (base) & $0.1$ & $0.2$ & $0.3$ & $0.4$ & $0.5$ & $0.6$ & $0.7$ & $0.8$ & $0.9$ & $1.0$ & $1.1$ & $1.3$ & $1.4$ & $1.5$ \\
    \midrule
    Qwen2.5-7B $\rightarrow$ Dream-7B & IFEval
    & 50.5 & --- & 54.1 & --- & --- & 58.4 & 58.7 & 59.1 & 58.4 & \sel{\textbf{59.3}} & 56.7& --- & --- & --- & --- \\
    Qwen2.5-Coder $\rightarrow$ Dream-Coder & HumanEval+
    & 57.9 & --- & --- & --- & 65.9 & 65.9 & \sel{\textbf{67.1}} & 61.0 & 59.1 & --- & ---& --- & --- & --- & --- \\
    Qwen2.5-Coder $\rightarrow$ DiffuCoder & HumanEval+
     & 61.6 & \sel{\textbf{60.4}} & 57.3 & --- & 47.6 & --- & --- & --- & --- & --- & ---& --- & --- & --- & --- \\
    AceCoder-Rule $\rightarrow$ DiffuCoder (RL) & MBPP+
     & 65.9 & 66.1 & \sel{\textbf{66.9}} & 65.3 & 65.3 & --- & --- & --- & --- & --- & ---& --- & --- & --- & --- \\
    Qwen3-8B $\rightarrow$ DreamReasoner & MATH-500
     & --- & --- & --- & --- & 74.0 & \sel{\textbf{76.0}} & 75.8 & --- & --- & --- & ---& --- & --- & --- & --- \\
    Ministral-3-3B-Instruct $\rightarrow$ Nemotron-LD-3B & IFEval
    & 25.2 & --- & --- & --- & --- & --- & ---
    & 53.4 & 56.2 & \sel{\textbf{59.6}} & 58.1
    & --- & --- & --- & --- \\
    Ministral-3-8B-Instruct $\rightarrow$ Nemotron-LD-8B & IFEval
     & 28.2 & --- & --- & --- & --- & --- & --- & --- & --- & 59.0 & 61.6 & 61.0 & 62.2 & \sel{\textbf{64.9}} & 63.2 \\
    \addlinespace[1pt]
    Ministral-3-8B-Reasoning $\rightarrow$ Nemotron-LD-8B & IFEval
     & 28.2 & --- & --- & --- & 53.5 & --- & 55.1 & 56.9 & \sel{\textbf{57.6}} & 57.6 & 56.8 & --- & --- & --- & --- \\
    Gemma~4 $\rightarrow$ DiffusionGemma & SLAKE
    & 63.2 & --- & 68.0 & --- & 71.6 & --- & 74.9 & --- & 76.8 & 76.9
    & \sel{\textbf{77.1}} & 75.5 & --- & --- & --- \\
    Gemma~4 $\rightarrow$ DiffusionGemma & VQA-RAD
    & 57.4 & --- & 65.0 & --- & 68.5 & 72.9 & 72.3 & \sel{\textbf{74.1}} & 71.8 & ---
    & 70.5 & --- & --- & --- & --- \\
    
    \bottomrule
    \end{tabular}}
\end{table*}

\begin{table*}[ht]
\footnotesize\centering
\setlength{\tabcolsep}{4pt}\renewcommand{\arraystretch}{1.1}
\caption{$D_p+\alpha\tau_A$: the AR vector added to the post-trained diffusion model. Shading and bold as above. All results are in percentages.
}
\label{tab:sweep-addition}
\resizebox{0.98\linewidth}{!}{%
\begin{tabular}{@{}l l ccccccccccccc@{}}
\toprule
AR $\rightarrow$ Diffusion & Anchor & \multicolumn{13}{c@{}}{$\alpha$} \\
\cmidrule(l){3-15}
 &  & $0$ (parent) & $0.1$ & $0.2$ & $0.3$ & $0.4$ & $0.5$ & $0.6$ & $0.7$ & $0.8$ & $0.9$ & $1.0$ & $1.1$ & $1.2$ \\
\midrule
Qwen2.5-7B $\rightarrow$ Dream-7B & IFEval
 & 63.0 & --- & 67.7 & --- & 70.9 & --- & 71.0 & 73.3 & 73.5 & 74.5 & 74.9 & \sel{\textbf{77.1}} & 75.5 \\
Qwen2.5-Coder $\rightarrow$ Dream-Coder & HumanEval+
 & 77.4 & --- & 75.0 & 76.8 & \sel{\textbf{77.4}} & \textbf{77.4} & 75.6 & --- & --- & --- & --- & --- & --- \\
Qwen2.5-Coder $\rightarrow$ DiffuCoder & HumanEval+
 & 65.9 & --- & 66.5 & --- & 68.9 & \sel{\textbf{69.5}} & 67.7 & 60.4 & --- & --- & --- & --- & --- \\
AceCoder-Rule $\rightarrow$ DiffuCoder (RL) & MBPP+
 & 66.9 & --- & --- & --- & 66.9 & 66.9 & 66.4 & --- & 66.9 & --- & --- & --- & --- \\
Qwen3-8B $\rightarrow$ DreamReasoner & MATH-500
 & 93.6 & --- & 93.6 & 93.6 & 93.8 & \sel{\textbf{94.4}} & 93.6 & 94.2 & --- & --- & --- & --- & --- \\
 Ministral-3-3B-Instruct $\rightarrow$ Nemotron-LD-3B & IFEval
& 68.0 & 68.1 & 69.1 & 69.2 & \sel{\textbf{71.0}} & 70.0
& 69.8 & 69.4 & 66.5 & 64.6 & --- & --- & --- \\
Ministral-3-8B-Instruct $\rightarrow$ Nemotron-LD-8B & IFEval
 & 69.8 & 69.6 & 71.3 & 73.5 & 74.3 & \sel{\textbf{75.4}} & 72.9 & 73.2 & 73.6 & 71.4 & --- & --- & --- \\
\addlinespace[1pt]
Ministral-3-8B-Reasoning $\rightarrow$ Nemotron-LD-8B & MATH-500
 & 87.2 & 88.4 & 90.8 & \sel{\textbf{91.6}} & 89.2 & --- & 79.6 & --- & 35.4 & --- & --- & --- & --- \\
Gemma~4 $\rightarrow$ DiffusionGemma & SLAKE
& 83.1 & \sel{\textbf{83.4}} & 82.8 & --- & 82.4 & --- & 81.6
& --- & 79.3 & --- & 77.9 & --- & --- \\

Gemma~4 $\rightarrow$ DiffusionGemma & VQA-RAD
& 71.6 & --- & 73.4 & 73.4 & \sel{\textbf{74.7}} & 74.3 & 73.2
& --- & 74.3 & --- & 71.4 & --- & --- \\

\bottomrule
\end{tabular}}

\end{table*}

\begin{table*}[ht]
\footnotesize\centering
\setlength{\tabcolsep}{5pt}\renewcommand{\arraystretch}{1.1}
\caption{$D_b+\alpha\tau_D$: the model's own post-training vector rescaled, no AR term. Shading and bold as above. All results are in percentages.
}
\label{tab:sweep-rescale}
\resizebox{0.98\linewidth}{!}{%
\begin{tabular}{@{}l l cccccccccc@{}}
\toprule
Diffusion (no AR term) & Anchor & \multicolumn{10}{c@{}}{$\alpha$} \\
\cmidrule(l){3-12}
& & $0.1$ & $0.2$ & $0.3$ & $0.4$ & $0.5$ & $0.6$ & $0.7$ & $0.8$ & $0.9$ & $1.0$ (parent) \\
\midrule
Qwen2.5-7B $\rightarrow$ Dream-7B & IFEval
& --- & --- & --- & --- & --- & 61.5 & --- & 61.3 & \sel{\textbf{63.4}} & 63.0 \\
Qwen2.5-Coder $\rightarrow$ Dream-Coder & HumanEval+
& --- & --- & --- & --- & --- & --- & --- & 72.0 & \sel{\textbf{78.7}} & 77.4 \\
Qwen2.5-Coder $\rightarrow$ DiffuCoder & HumanEval+
& --- & --- & --- & 68.3 & 68.9 & \sel{\textbf{70.1}} & 65.9 & 66.5 & 65.9 & 65.9 \\
AceCoder-Rule $\rightarrow$ DiffuCoder (RL) & MBPP+
& --- & 65.6 & --- & 66.7 & 69.3 & 68.8 & \sel{\textbf{69.6}} & 67.5 & --- & 66.9 \\
Qwen3-8B $\rightarrow$ DreamReasoner & MATH-500
& --- & --- & --- & --- & 88.4 & 89.0 & 88.8 & \sel{\textbf{93.4}} & 89.8 & 93.6 \\
Ministral-3-3B-Instruct $\rightarrow$ Nemotron-LD-3B & IFEval
& 28.4 & 49.7 & 60.9 & 67.3 & 70.1 & 70.3
& \sel{\textbf{72.0}} & 71.3 & 70.5 & 68.0 \\
Ministral-3-8B-Instruct $\rightarrow$ Nemotron-LD-8B & IFEval
& 20.8 & 21.4 & 24.2 & 56.4 & 70.1 & \sel{\textbf{71.7}} & 70.8 & 70.5 & 71.5 & 69.8 \\
\addlinespace[1pt]
Ministral-3-8B-Reasoning $\rightarrow$ Nemotron-LD-8B & MATH-500
& --- & 0.0 & --- & 41.4 & --- & 83.6 & 87.4 & 90.2 & \sel{\textbf{90.8}} & 87.2 \\
Gemma~4 $\rightarrow$ DiffusionGemma & SLAKE
& --- & 68.7 & --- & 75.6 & --- & 80.3 & 82.6
& \sel{\textbf{84.0}} & 83.4 & 83.1 \\

Gemma~4 $\rightarrow$ DiffusionGemma & VQA-RAD
& --- & 66.7 & --- & 69.0 & --- & 71.2 & ---
& 71.4 & \sel{\textbf{72.5}} & 71.6 \\

\bottomrule
\end{tabular}}

\end{table*}

\paragraph{Efficiency of hyperparameter search}

\providecommand{\methodname}{A2D}
\begin{table}[ht]
\centering\footnotesize
\setlength{\tabcolsep}{5pt}
\renewcommand{\arraystretch}{1.1}
\caption{\textbf{Short-budget IFEval comparison on the 8B \methodname{}
merge.} Both budgets evaluate the same 541 prompts with thinking
off and select $\lambda=0.4$. Scores are percentages; median lengths
are generated token counts. Bold marks the best score per budget.}
\label{tab:ifeval-budget}
\begin{tabular}{@{}c rr rr@{}}
\toprule
& \multicolumn{2}{c}{1,280-token budget}
& \multicolumn{2}{c@{}}{8,192-token budget} \\
\cmidrule(lr){2-3}\cmidrule(l){4-5}
$\lambda$ & Score & Median length & Score & Median length \\
\midrule
0.3 & 74.03 & 314 & 74.07 & 308 \\
\rowcolor{blue!5}
0.4 & \textbf{78.00} & 332 & \textbf{78.76} & 336 \\
0.5 & 75.59 & 323 & 76.20 & 315 \\
\bottomrule
\end{tabular}
\end{table}

Evaluating the transfer hyperparameter sweeps for the Nemotron-LD variants in Table~\ref{tab:sweep-transfer} on all 541 IFEval prompts with the full $8{,}192$-token generation budget was too expensive, so we used a $1{,}280$-token budget while retaining the full dataset for coefficient selection. This choice was motivated by a separate budget comparison on the Nemotron-LD-8B protected merge (Table~\ref{tab:ifeval-budget}): the shorter budget preserved the ranking and selected coefficient as well as median output lengths across the tested points.

\clearpage
\section{Additional Results}
\label{app:additional_res}

In this section, we report additional experimental results.

\subsection{Full results for analyses and ablations}

Here we put the full results for the analyses and ablations conducted in \autoref{sec:analysis}.

\begin{figure}[ht]
    \centering

    \includegraphics[width=0.47\linewidth]
    {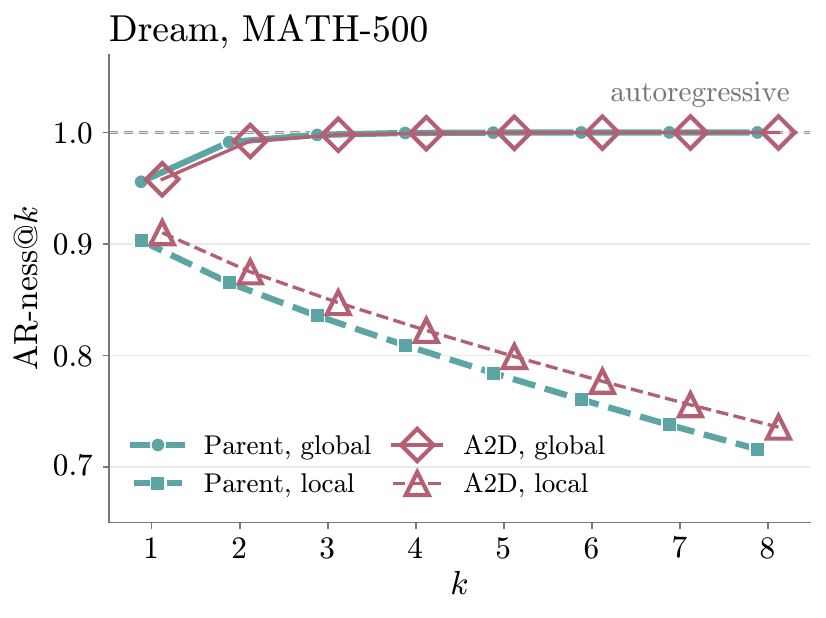}
    \hfill
    \includegraphics[width=0.47\linewidth]
    {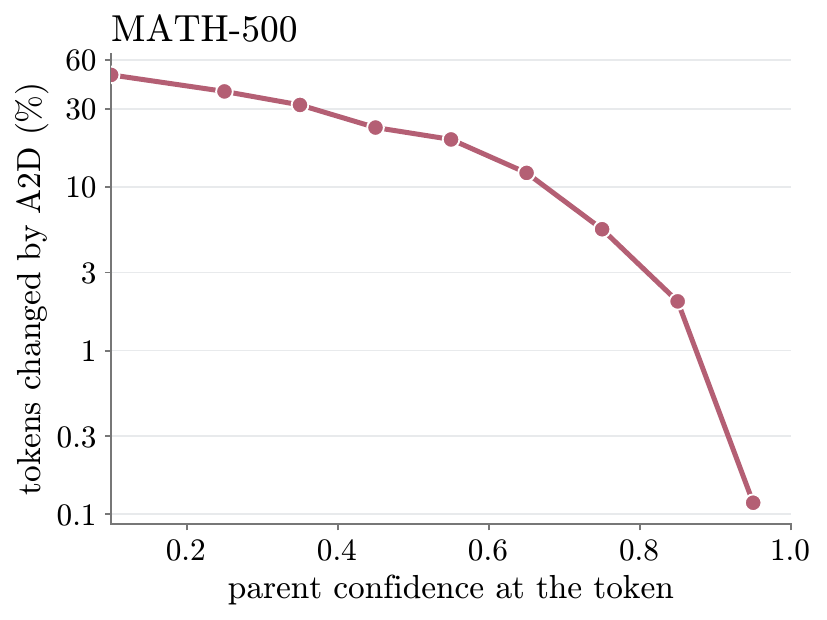}

    \caption{\textbf{How A2D alters diffusion generation.}
    \textbf{Left:} A2D largely preserves the parent's decoding order.
    \textbf{Right:} Prediction changes concentrate on low-confidence tokens.}
    \label{fig:generation_behavior}

    \vspace{-6pt}
\end{figure}

\begin{itemize}
    \item \autoref{tab:gain-decomp} presents the full results for the decomposition of the contributions by adding the AR task vector and downscaling the diffusion task vector.

    \item \autoref{tab:reverse-main} presents the full results for transfering the diffusion vector back to the AR base model.

    \item \autoref{fig:generation_behavior} (a) compares the "AR-ness" between the diffusion parent model and the A2D merged model, showing that A2D does not shift the model systematically toward autoregressive generation. 
    \item \autoref{fig:generation_behavior} (b) shows that A2D mainly changes the diffusion parent's token prediction at positions where the diffusion parent has a low confidence.
    \item \autoref{tab:module_ablation} presents the module-wise ablation of the
    transferred AR update. For \methodname-Transfer, removing the MLP component causes a
    substantial performance drop, while removing attention has little effect.
    For \methodname-Merge, reverting the MLP modules to the diffusion parent nearly preserves
    full performance, whereas reverting attention causes a larger drop, suggesting
    that MLPs mainly carry standalone AR transfer while attention provides more
    complementary gains during merging.
    \item \autoref{tab:continued} shows that continued-pretraining vectors do not transfer effectively, suggesting that \methodname is most suitable for relatively local post-training updates rather than pretraining-scale model transformations.
\end{itemize}

\providecommand{\tbd}{{\color{red}\textbf{??}}}
\begin{table}[ht]
    \centering\footnotesize
    \setlength{\tabcolsep}{3pt}
    \renewcommand{\arraystretch}{1.1}
    \caption{\textbf{Sources of the A2D gain.} Best-performing point for each construction, with coefficients selected on the anchor task. Avg.\ rank is the mean within-pair rank (1 = best) and Avg.\ Acc.\ is the mean over the seven pairs. All results are in percentages.}
    \label{tab:gain-decomp}
    \resizebox{\linewidth}{!}{%
    \begin{tabular}{@{}lccccccc!{\color{black!30}\vrule}cc@{}}
    \toprule
    & Dream & DreamCoder & DiffuCoder & DiffuCoder-RL & DreamReasoner & Nemotron-3B
    & DiffusionGemma
    & \multirow{2}{*}{Avg.\ rank$\downarrow$}
    & \multirow{2}{*}{Avg.\ Acc.} \\
    & \scriptsize IFEval
    & \scriptsize HumanEval+
    & \scriptsize HumanEval+
    & \scriptsize MBPP+
    & \scriptsize MATH-500
    & \scriptsize IFEval
    & \scriptsize VQA-RAD
    & & \\
    \midrule
    \rowcolor{gray!5}
    Diffusion parent $\ D_p$
    & 63.0 & 77.4 & 65.9 & 66.9 & 93.6 & 68.0 & 71.6 & 3.7 & 72.3 \\
    
    $+\tau_A$ only $\ D_p+\lambda\tau_A$
    & \textbf{77.1} & 77.4 & 69.5 & 66.9 & 94.4 & 71.0 & \textbf{74.7} & 2.4 & 75.9 \\
    
    $-\tau_D$ only $\ D_p-\lambda\tau_D$
    & 63.4 & 78.7 & 70.1 & \textbf{69.6} & 93.4 & 72.0 & 72.5 & 2.4 & 74.2 \\
    
    \rowcolor{blue!5}
    \methodname\-merge $\ D_p+\lambda(\tau_A-\tau_D)$
    & 72.8 & \textbf{79.3} & \textbf{74.4} & 69.0 & \textbf{94.8} & \textbf{72.9} & 74.5
    & \textbf{1.4} & \textbf{76.8} \\
    \bottomrule
    \end{tabular}}
    \end{table}

\providecommand{\tbd}{{\color{red}\textbf{??}}}
\begin{table*}[ht]
\centering\footnotesize
\setlength{\tabcolsep}{3pt}
\renewcommand{\arraystretch}{1.1}
\caption{\textbf{Reverse direction: diffusion post-training vector into the AR model.} AR's rows use their native budget.
All results are in percentages.}
\resizebox{0.98\linewidth}{!}{%
\begin{tabular}{@{}lcccc!{\color{black!30}\vrule}cc!{\color{black!30}\vrule}ccc@{}}
\toprule
& \multicolumn{4}{c!{\color{black!30}\vrule}}{Dream-7B$\rightarrow$Qwen2.5-7B}
& \multicolumn{2}{c!{\color{black!30}\vrule}}{DiffuCoder$\rightarrow$Qwen2.5-Coder}
& \multicolumn{3}{c@{}}{Nemotron-LD-3B$\rightarrow$Ministral-3-3B} \\
\cmidrule(lr){2-5}\cmidrule(lr){6-7}\cmidrule(l){8-10}
& \colorbox{green!5}{IFEval} & IFBench & GSM8K & MATH-500
& \colorbox{green!5}{HumanEval+} & MBPP+
& \colorbox{green!5}{IFEval} & MATH-500 & MBPP \\
\midrule
\rowcolor{gray!5}
Diffusion donor
& 63.0 & 19.0 & 83.0 & 38.4 & 65.9 & 65.9
& 68.0 & 76.8 & 73.0 \\
\midrule
AR base
& 40.5 & 16.3 & \textbf{62.5} & \textbf{17.4} & 74.4 & 62.2
& 45.4 & \textbf{60.2} & \textbf{64.3} \\
\rowcolor{blue!5}
\methodname\ (transfer, $\alpha{=}1.0$ / $0.1$ / $0.9$)
& \textbf{46.7} & \textbf{19.0} & 56.0 & 16.2 & 74.4 & \textbf{64.8}
& \textbf{55.7} & 35.0 & 55.8 \\
\midrule
AR instruct
& 75.3 & \textbf{31.7} & \textbf{83.0} & 56.0 & 86.6 & 71.7
& 67.0 & \textbf{82.4} & 76.2 \\
\rowcolor{blue!5}
\methodname\ (merge, $\lambda{=}0.1$ / $0.2$ / $0.2$)
& \textbf{77.3} & 30.7 & 82.9 & \textbf{56.4} & 86.6 & 71.7
& \textbf{69.0} & 81.4 & \textbf{77.0} \\
\bottomrule
\end{tabular}}
\label{tab:reverse-main}
\end{table*}

\begin{table}[t]
    \centering
    \footnotesize
    \setlength{\tabcolsep}{6pt}
    \renewcommand{\arraystretch}{1.08}
    \caption{\textbf{Module-wise ablation of the transferred AR update.}
    MATH-500 accuracy on Dream-7B.
    For A2D-Merge, the ablated module is reverted to the diffusion parent
    $D_p$, while all remaining modules use the merged weights. All results are in percentages.}
    \label{tab:module_ablation}
    \begin{tabular}{@{}lcc@{}}
        \toprule
        & A2D-Transfer & A2D-Merge \\
        \midrule
        Full model       & \textbf{40.8} & \textbf{45.8} \\
        w/o AR attention & 40.8 & 43.0 \\
        w/o AR MLP       & 30.4 & 45.6 \\
        \bottomrule
    \end{tabular}
\end{table}

\providecommand{\tbd}{{\color{red}\textbf{??}}}
\begin{table}[ht]
\footnotesize
\centering
\setlength{\tabcolsep}{4pt}
\renewcommand{\arraystretch}{1.1}
\caption{\textbf{Continued pre-training vectors do not transfer.}
Adding the AR continued-pretraining vector to the diffusion base does not improve performance, even at small coefficients. These vectors are $70$--$82\times$ larger than the post-training vectors used in our main experiments. \textbf{Bold} denotes the best $\alpha{>}0$ result; all results are percentages.}
\label{tab:continued}

\vspace{2pt}

\resizebox{\linewidth}{!}{%
\begin{tabular}{@{}l l l >{\columncolor{gray!8}}c ccc@{}}
\toprule
 & Recipient & Benchmark & \multicolumn{4}{c@{}}{$\alpha$ in $D_b+\alpha\tau_{\mathrm{CPT}}$} \\
\cmidrule(l){4-7}
& & & $0$ & $0.01$ & $0.05$ & $1.0$ \\
\midrule
Math ($82\times$), Qwen2.5-Math $-$ Qwen2.5 & Dream-7B-Base & GSM8K & 76.6 & \textbf{76.6} & 74.8 & 0.0 \\
Code ($70\times$), Qwen2.5-Coder $-$ Qwen2.5 & Dream-Coder-Base & HumanEval & 63.4 & \textbf{61.6} & 57.3 & 0.0 \\
\bottomrule
\end{tabular}}

\end{table}

\subsection{Full results of Dream-7B}

Here, we report the results used in ~\autoref{fig:method_name_radar_base} and in ~\autoref{fig:method_name_radar_post} in Table~\ref{tab:dreamv0-full}.

\providecommand{\tbd}{{\color{red}\textbf{??}}}
\begin{table}[ht]
\footnotesize\centering
\setlength{\tabcolsep}{5pt}\renewcommand{\arraystretch}{1.1}
\caption{\textbf{Qwen2.5-7B $\rightarrow$ Dream-7B on six tasks.} One transfer model ($\alpha{=}0.9$) and one merged
model ($\lambda{=}0.6$), both selected on IFEval, improve every task over the corresponding baseline. All results are in percentages.}
\label{tab:dreamv0-full}
\begin{tabular}{@{}lcccccc@{}}
\toprule
& \multicolumn{6}{c@{}}{Qwen2.5-7B $\rightarrow$ Dream-7B} \\
\cmidrule(l){2-7}
& IFEval & IFBench & GSM8K & MATH-500 & HumanEval+ & MBPP+ \\
\midrule
\rowcolor{gray!5}
AR donor & 75.3 & 31.7 & 83.0 & 36.6 & 76.8 & 69.0 \\
\midrule
Base & 49.7 & 14.0 & 66.5 & 20.8 & 48.8 & 59.3 \\
\rowcolor{blue!5}
\methodname\ (transfer) & \textbf{59.2} & \textbf{25.0} & \textbf{81.7} & \textbf{40.2} & \textbf{54.9} & \textbf{62.7} \\
\midrule
Instruct & 63.0 & 19.0 & 83.0 & 38.4 & 51.8 & 57.1 \\
\rowcolor{blue!5}
\methodname\ (merge) & \textbf{72.8} & \textbf{22.3} & \textbf{83.7} & \textbf{45.0} & \textbf{59.8} & \textbf{58.7} \\
\bottomrule
\end{tabular}
\end{table}

\subsection{Full results of Nemotron-LD-3B}
\label{app:nemontron3B}

\providecommand{\methodname}{A2D}
\begin{table*}[h]
\footnotesize\centering
\setlength{\tabcolsep}{4pt}
\renewcommand{\arraystretch}{1.1}
\caption{\textbf{  Ministral-3-3B-Instruct $\rightarrow$ Nemotron-LD-3B.}
DLM and AR denote diffusion and autoregressive decoding of the same checkpoint.
Mean excludes IFEval. \textbf{Bold} marks the better result within each
Instruct-merge pair. All results are in percentages.}
\label{tab:nemotron3b}

\resizebox{0.99\linewidth}{!}{%
\begin{tabular}{@{}llcccccccccc|c@{}}
\toprule
& Mode
& GSM8K & MATH-500 & HumanEval & MBPP
& \colorbox{green!5}{IFEval}
& MMLU & GPQA & AIME24 & AIME25 & LCB-CPP
& Mean \\
\midrule

\rowcolor{gray!5}
AR donor & AR
& 91.3 & 83.0 & 80.5 & 75.9 & 65.1
& 67.9 & 40.9 & 30.0 & 13.3 & 27.1 & 56.7 \\

\midrule

Base (raw protocol) & DLM
& 59.1 & 40.0 & 64.6 & 70.9 & 23.2
& 62.0 & \textbf{35.4} & 0.0 & 0.0 & 9.9 & 38.0 \\

Base (Instruct protocol) & DLM
& 78.0 & 34.6 & 39.6 & 38.6 & 25.8
& 55.1 & 26.3 & 0.0 & 0.0 & 4.0 & 30.7 \\

\rowcolor{blue!5}
\methodname\ (transfer) & DLM
& \textbf{88.1} & \textbf{52.6} & \textbf{72.6} & \textbf{74.3} & \textbf{60.6}
& \textbf{62.8} & 33.3 & \textbf{3.3} & \textbf{3.3} & \textbf{14.1} & \textbf{44.9} \\

Base (raw protocol) & AR
& 52.1 & 50.4 & 71.3 & 72.8 & 21.1
& 58.0 & 32.8 & \textbf{10.0} & \textbf{13.3}
& \textbf{14.5} & 41.7 \\

\rowcolor{blue!5}
\methodname\ (transfer) & AR
& \textbf{89.2} & \textbf{60.4} & \textbf{73.8}
& \textbf{74.6} & \textbf{59.0}
& \textbf{62.6} & \textbf{33.3} & 6.7 & 0.0
& \textbf{14.5} & \textbf{46.1} \\
\midrule

Instruct & DLM
& 89.2 & 76.8 & 73.2 & 73.0 & 68.0
& 72.1 & 35.4 & 10.0 & \textbf{13.3} & 16.3 & 51.0 \\

\rowcolor{blue!5}
\methodname\ (merge) & DLM
& \textbf{90.1} & \textbf{79.6} & \textbf{75.0}
& \textbf{75.7} & \textbf{72.9} & \textbf{72.9}
& \textbf{43.4} & \textbf{16.7} & \textbf{13.3}
& \textbf{18.9} & \textbf{54.0} \\

\addlinespace[2pt]

Instruct & AR
& 87.6 & 76.0 & 73.8 & 69.8 & 68.8
& 71.9 & \textbf{41.9} & \textbf{23.3}
& \textbf{13.3} & 17.6 & 52.8 \\

\rowcolor{blue!5}
\methodname\ (merge) & AR
& \textbf{89.3} & \textbf{81.0} & \textbf{79.9}
& \textbf{75.1} & \textbf{72.8} & \textbf{72.5}
& 38.4 & 20.0 & \textbf{13.3}
& \textbf{18.7} & \textbf{54.3} \\

\bottomrule
\end{tabular}}
\end{table*}

\autoref{tab:nemotron3b} reports the full Nemotron-LD-3B results under both
diffusion and autoregressive decoding.
For the diffusion Base, we evaluate two prompting protocols.
The main paper reports the native raw-prompt result, which provides the stronger overall Base performance.
For completeness, we additionally evaluate the Base using the same instruction
prompt format as A2D-Transfer, providing a prompt-matched comparison.

Under either protocol, A2D-Transfer substantially strengthens the Base
checkpoint, bringing it close to or beyond the Instruct model on several
benchmarks.
A2D-Merge then provides further improvements over the Instruct checkpoint under
both diffusion and autoregressive decoding.

\subsection{Results of Nemotron-LD-8B}
\label{app:nemontron8b}

\providecommand{\tbd}{{\color{red}\textbf{??}}}
\providecommand{\methodname}{A2D}
\begin{table*}[h]
\footnotesize\centering
\setlength{\tabcolsep}{4pt}\renewcommand{\arraystretch}{1.1}
\caption{\textbf{Ministral-3-8B $\rightarrow$ Nemotron-LD-8B.}
DLM and AR denote diffusion and autoregressive decoding of the same checkpoint.
Mean excludes IFEval and, when unavailable, MMLU, which under some settings exhibits repetitive generation and requires over four days to complete.
\textbf{Bold} marks the better result within each corresponding pair. All results are in percentages.}
\label{tab:nemotron8b}

\resizebox{0.99\linewidth}{!}{%
\begin{tabular}{@{}llcccccccccc|c@{}}
\toprule
& Mode & GSM8K & MATH-500 & HumanEval & MBPP
& \colorbox{green!5}{IFEval} & MMLU & GPQA
& AIME24 & AIME25 & LCB-CPP & Mean \\
\midrule

\rowcolor{gray!5}
AR donor & AR
& 93.2 & 87.8 & 77.4 & 77.8 & 69.5 & 76.7
& 50.0 & 30.0 & 26.7 & 34.8 & 61.6 \\

\midrule

Base (raw protocol) & DLM & 64.7 & 51.2 & \textbf{77.4} & \textbf{82.0} & 28.2 & 38.4 & 34.3 & 13.3 & \textbf{20.0} & \textbf{17.0} & 44.3 \\
Base (Instruct protocol) & DLM & 78.2 & 59.8 & 4.9 & 13.2 & 21.5 & --- & 9.1 & 6.7 & 10.0 & 2.2 & \textit{23.0} \\
\rowcolor{blue!5}
\methodname\ (transfer) & DLM & \textbf{89.5} & \textbf{71.6} & 75.0 & 81.7 & \textbf{65.7} & 72.4 & \textbf{37.4} & 13.3 & 6.7 & 11.2 & \textbf{51.0} \\
Base (raw protocol) & AR & 63.8 & 60.2 & \textbf{76.2} & \textbf{79.9} & 26.0 & 53.6 & 33.3 & \textbf{13.3} & \textbf{20.0} & \textbf{13.4} & 46.0 \\
\rowcolor{blue!5}
\methodname\ (transfer) & AR & \textbf{89.7} & \textbf{69.8} & 74.4 & 79.4 & \textbf{62.6} & \textbf{70.1} & \textbf{39.9} & 6.7 & 13.3 & 12.8 & \textbf{50.7} \\

\midrule

Instruct & DLM
& 93.7 & \textbf{89.2} & \textbf{73.8} & 62.2
& 69.8 & \textbf{79.1} & \textbf{44.4}
& 30.0 & \textbf{30.0} & \textbf{29.3} & \textbf{59.1} \\

\rowcolor{blue!5}
\methodname\ (merge) & DLM
& \textbf{94.5} & 89.0 & 71.3 & \textbf{65.3}
& \textbf{78.8} & 78.6 & 42.9
& \textbf{33.3} & 26.7 & 28.0 & 58.8 \\

Instruct & AR
& 93.5 & 88.6 & 78.0 & 65.1 & 70.8
& 79.8 & \textbf{48.0} & \textbf{33.3}
& 26.7 & \textbf{35.5} & \textbf{60.9} \\

\rowcolor{blue!5}
\methodname\ (merge) & AR
& \textbf{93.9} & \textbf{89.4} & \textbf{79.3} & \textbf{72.2}
& \textbf{74.9} & 78.8 & 47.0
& 30.0 & 26.7 & 31.1 & \textbf{60.9} \\

\midrule

\multicolumn{13}{@{}l}{\emph{thinking}, official budget ($8{,}192$ total / $6{,}000$ thinking), reasoning donor}\\

Instruct & DLM
& 92.4 & 87.2 & \textbf{86.0} & \textbf{87.3}
& \textbf{75.5} & --- & 36.4
& 33.3 & 30.0 & 26.9 & 59.9 \\

\rowcolor{blue!5}
\methodname\ (merge) & DLM
& \textbf{94.1} & \textbf{92.6} & 79.3 & 86.2
& 74.7 & --- & \textbf{44.4}
& \textbf{46.7} & \textbf{33.3} & \textbf{38.1} & \textbf{64.3} \\

\midrule

\multicolumn{13}{@{}l}{\emph{thinking, double budget} ($16{,}384$ total / $12{,}000$ thinking), reasoning donor}\\

Nemotron-LD-8B-Instruct & DLM
& 92.2 & 87.0 & \textbf{85.4} & \textbf{87.3}
& \textbf{75.6} & --- & 40.4
& 33.3 & 30.0 & 27.3 & 60.4 \\

\rowcolor{blue!5}
\methodname\ (merge) & DLM
& \textbf{94.0} & \textbf{92.2} & 79.3 & 86.5
& 74.6 & --- & \textbf{46.0}
& \textbf{53.3} & \textbf{33.3} & \textbf{40.5} & \textbf{65.6} \\

\bottomrule
\end{tabular}}
\end{table*}

Here, we provide the full results for Nemotron-LD-8B in ~\autoref{tab:nemotron8b}. 
At the 8B scale, we evaluate A2D with both instruction-tuning and
reasoning-oriented AR updates.

\begin{figure}[h]
\centering
\includegraphics[width=\textwidth]{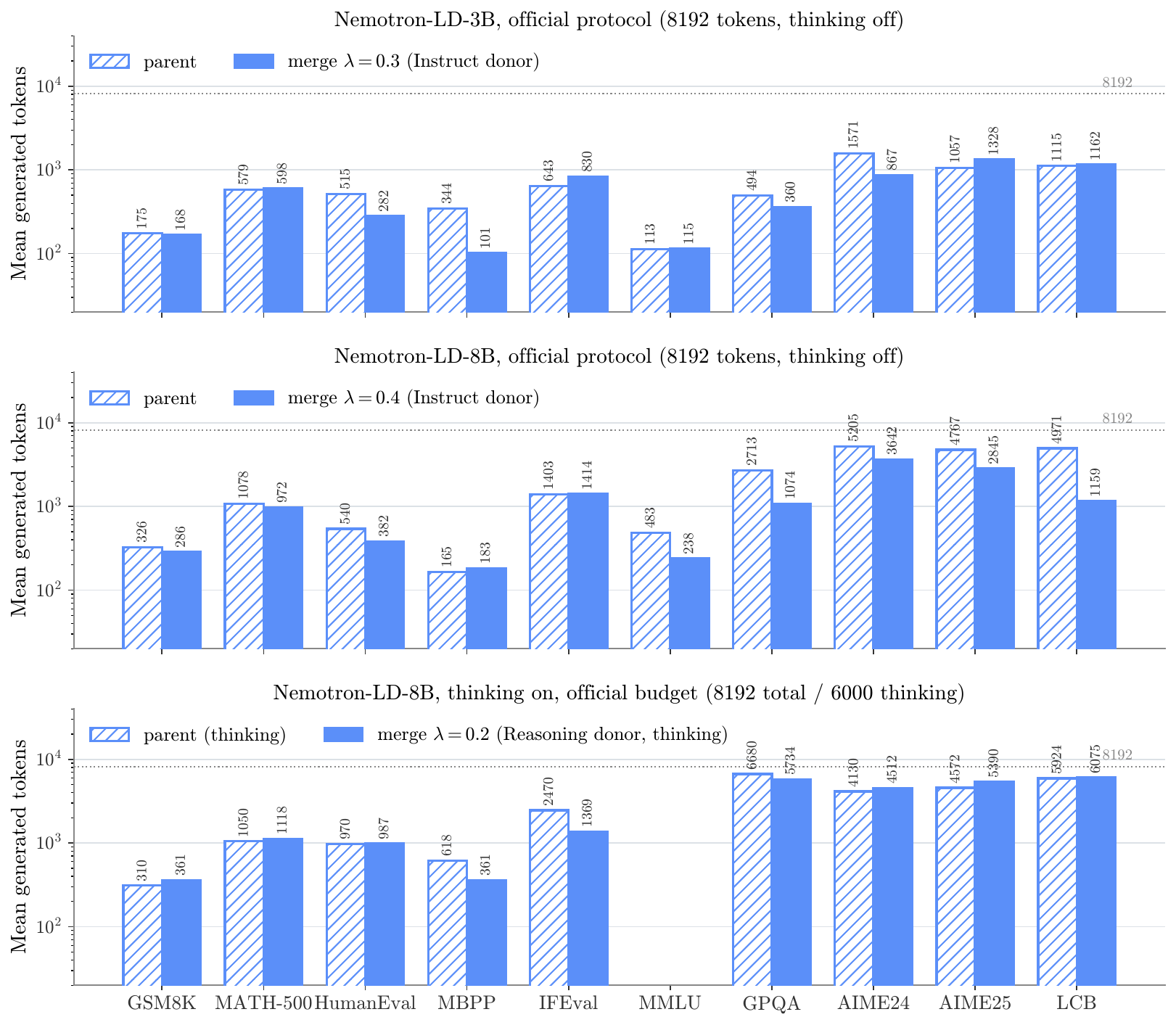}
\caption{\textbf{Generation-length mismatch in Nemotron-LD.}
The 8B model, unlike the 3B one, exhibits longer, reasoning-like generations even with thinking disabled. Enabling thinking substantially reduces the length mismatch between the parent and merged models.}
\label{fig:nemo-len}
\end{figure}

\paragraph{Instruction-tuning updates.}
With the Ministral-3-8B-Instruct donor, A2D-Transfer broadly improves the
diffusion base, while A2D-Merge provides gains on several benchmarks under both
diffusion and AR decoding.
The merging gains are less consistent than for the 3B model.
Importantly, this setting exhibits a mismatch in generation behavior:
Nemotron-LD-8B produces long, reasoning-like responses on reasoning-intensive
tasks even with thinking disabled, whereas merging the Instruct update
substantially shortens these generations
(\autoref{fig:nemo-len}).
Thus, comparisons on these tasks are partially confounded by changes in
effective reasoning behavior in addition to the weight-space composition itself.

\paragraph{Reasoning updates.}
Separately, we evaluate A2D using the Ministral-3-8B-Reasoning donor, with
thinking enabled for both the diffusion parent and the merged model.
Unlike the instruction-tuning setting above, both models use the same explicit
thinking mode, avoiding the generation-behavior mismatch.
Under the official generation budget, A2D-Merge improves the mean score from
$59.9$ to $64.3$, with substantial gains on reasoning-related benchmarks such as MATH-500
($87.2\!\rightarrow\!92.6$), GPQA ($36.4\!\rightarrow\!44.4$),
AIME24 ($33.3\!\rightarrow\!46.7$), and LiveCodeBench
($26.9\!\rightarrow\!38.1$).
The gains remain strong when doubling the generation budget, where the mean
further increases from $60.4$ to $65.6$.
These results show that reasoning-oriented AR post-training updates can be
effectively composed with diffusion models at the 8B scale under matched
reasoning behavior.

\subsection{DiffusionGemma with SLAKE Post-training}
\begin{table}[ht]
    \footnotesize
    \centering
    \setlength{\tabcolsep}{6pt}
    \renewcommand{\arraystretch}{1.1}
    \setlength{\fboxsep}{1.5pt}

    \caption{\textbf{AR SLAKE post-training transfer from Gemma to DiffusionGemma.}
    \colorbox{gray!5}{AR donor} denotes the AR checkpoint providing the task vector.
    Base denotes the instruction-tuned diffusion checkpoint.
    \colorbox{green!5}{Anchor task} used for hyperparameter selection is highlighted.
    \textbf{Bold} marks the best result compared to the corresponding baseline. All results are in percentages.}
    \label{tab:diffugemma-slake-full}

    \begin{tabular}{@{}lccc@{}}
    \toprule
    & \multicolumn{3}{c}{  Gemma 4$\rightarrow$ DiffusionGemma} \\
    \cmidrule(l){2-4}
    & VQA-RAD
    & \colorbox{green!5}{SLAKE}
    & VQA-Med \\
    \midrule

    \rowcolor{gray!5}
    AR donor
    & 58.09 & 80.68 & 37.20 \\

    \midrule

    Base
    & \textbf{57.43} & 63.24 & \textbf{43.20} \\

    \rowcolor{blue!5}
    \methodname\ (transfer)
    & 56.54 & \textbf{77.10} & 37.20 \\

    \midrule

    Post-trained
    & 60.75 & 83.13 & 39.60 \\

    \rowcolor{blue!5}
    \methodname\ (merge)
    & \textbf{61.64} & \textbf{83.32} & \textbf{40.00} \\

    \bottomrule
    \end{tabular}
\end{table}

In the main paper, we report results for merging with DiffusionGemma using checkpoints fine-tuned on VQA-RAD. In Table~\ref{tab:diffugemma-slake-full}, we additionally report the same setup using checkpoints fine-tuned on SLAKE. Unlike on VQA-Med, where the AR and diffusion models achieve comparable performance, on SLAKE the AR model is substantially weaker than its diffusion counterpart, providing a complementary setting where the donor itself is less capable than the recipient.

Despite this gap, the AR SLAKE task vector transfers effectively to the diffusion base and further improves the diffusion post-trained model across all three benchmarks when transferred or merged. This shows that AR updates can remain complementary even when the donor itself is weaker than the recipient.

\subsection{Self-speculative decoding of Nemotron}

\providecommand{\methodname}{A2D}
\begin{table*}[ht]
\centering\footnotesize
\setlength{\tabcolsep}{4pt}
\caption{\textbf{Accuracy and efficiency of self-speculative decoding.}
TPF is $\sum_i T_i/\sum_i \mathrm{NFE}_i$, including thinking tokens.
Relative time is normalized generation time per token within each
dataset, model size, and thinking setting; lower is better.
Timings are observational and exclude model loading and final grading.
Accuracy results are percentages (MATH-500 accuracy; MBPP base-test
pass@1), while TPF and relative time are ratios. Accuracy is in percentages.}
\label{tab:nemotron-selfspec}
\resizebox{0.75\linewidth}{!}{%
\begin{tabular}{@{}l ccc rrr rrr@{}}
\toprule
& & & & \multicolumn{3}{c}{MATH-500}
& \multicolumn{3}{c@{}}{MBPP} \\
\cmidrule(lr){5-7}\cmidrule(l){8-10}
Model & & &
& Acc.\ $\uparrow$ & TPF $\uparrow$ & Rel.\ time $\downarrow$
& Acc.\ $\uparrow$ & TPF $\uparrow$ & Rel.\ time $\downarrow$ \\
\midrule
\multicolumn{10}{@{}l}{\emph{\footnotesize{Ministral-3-3B-Instruct $\rightarrow$
Nemotron-LD-3B, thinking off}}} \\
Instruct parent & & &
& 76.6 & 4.96 & 1.00 & 69.8 & 3.71 & 1.00 \\
\rowcolor{blue!5}
\methodname{} (merge) & & &
& \textbf{80.8} & \textbf{5.04} & \textbf{0.87}
& \textbf{74.9} & \textbf{3.90} & \textbf{0.80} \\
\midrule
\multicolumn{10}{@{}l}{\emph{\footnotesize{Ministral-3-8B-Instruct $\rightarrow$
Nemotron-LD-8B, thinking off}}} \\
Instruct parent & & &
& 87.0 & 5.34 & 1.00 & 68.0 & 3.42 & 1.00 \\
\rowcolor{blue!5}
\methodname{} (merge) & & &
& \textbf{88.6} & \textbf{6.07} & \textbf{0.88}
& \textbf{73.0} & \textbf{3.55} & \textbf{0.77} \\
\midrule
\multicolumn{10}{@{}l}{\emph{\footnotesize{Ministral-3-8B-Reasoning $\rightarrow$
Nemotron-LD-8B, thinking on}}} \\
Instruct parent & & &
& 87.6 & 3.70 & 1.00
& \textbf{87.8} & 3.77 & \textbf{1.00} \\
\rowcolor{blue!5}
\methodname{} (merge) & & &
& \textbf{91.4} & \textbf{4.04} & \textbf{0.81}
& 84.4 & \textbf{3.84} & 1.53 \\
\bottomrule
\end{tabular}}
\end{table*}

Table~\ref{tab:nemotron-selfspec} evaluates whether the benefits of \methodname{} extend to self-speculative decoding. We compare each merged checkpoint with its corresponding Instruct parent under matched decoding settings, reporting accuracy, tokens per forward evaluation (TPF), and generation time per token. We use NVIDIA's native linear self-speculative decoding, with diffusion drafting and autoregressive verification, a block size of 32, threshold 0, and no LoRA adapter. All runs use greedy decoding, with the same generation and thinking budgets
specified in Appendix~\ref{app:models_evals}.

Across the evaluated configurations, \methodname{} consistently increases TPF and generally improves accuracy. The non-thinking merges improve both accuracy and observed time per token on both benchmarks and at both model sizes. These benefits also extend to MATH-500 with the reasoning donor and thinking enabled. 
\subsection{Post-training types and norm ratios}
\label{app:norms}

\begin{table}[t]
\centering\footnotesize
\setlength{\tabcolsep}{4pt}\renewcommand{\arraystretch}{1.1}
\caption{\textbf{Geometry of post-training and paradigm-conversion directions across model pairs.}
Measurements use backbone parameters and, for Nemotron, compatible input embeddings after key mapping, excluding protected rows. Output heads are excluded throughout.}
\label{tab:pairs-geometry-full}
\resizebox{1.0\linewidth}{!}{%
\begin{tabular}{@{}l l rrrr cc@{}}
\toprule
\rowcolor{gray!5} 
diffusion $\leftarrow$ AR & post-training & $\|\tau_A\|$ & $\|\tau_D\|$ & $\|\Delta_b\|$ & $\|\Delta_p\|$ & $\cos(\Delta_b,\Delta_p)$ & $\cos(\tau_A,\tau_D)$ \\
\midrule
Qwen2.5-7B $\rightarrow$ Dream-7B 
& instruct (SFT)
& 23.0 & 31.5 & 729 & 732 & $+0.999$ & $+0.013$ \\

Qwen2.5-Coder$\rightarrow$  Dream-Coder 
& code (SFT)
& 87.0 & 84.6 & 589 & 602 & $+0.980$ & $+0.023$ \\

Qwen3-8B $\rightarrow$ DreamReasoner
& reasoning (SFT / SFT{+}RL)
& 189.2 & 295.8 & 459 & 556 & $+0.803$ & $+0.118$ \\

Ministral-3 $\rightarrow$ Nemotron-LD-3B 
& instruct (SFT), cross-arch.
& 43.1 & 17.1 & 280.9 & 271.4 & $+0.987$ & $+0.075$ \\

Ministral-3 $\rightarrow$ Nemotron-LD-8B 
& instruct (SFT), cross-arch.
& 42.2 & 16.3 & 387.0 & 380.4 & $+0.994$ & $+0.070$ \\
Qwen2.5-Coder $\rightarrow$ DiffuCoder    & code (SFT)                            & 87.0  & 33.3  & 627 & 630 & $+0.989$ & $+0.005$ \\
AceCoder-Rule $\rightarrow$ DiffuCoder   & code (RL) & 0.68  & 4.10  & 630 & 630 & $+1.000$ & $+0.004$ \\
\midrule
\rowcolor{gray!5}
\multicolumn{2}{@{}l}{\emph{Continued pre-training (knowledge) vectors, for contrast}}
& $\|v_A\|$ & $\|v_D\|$ & & & & $\cos(v_A,v_D)$ \\
\midrule
code knowledge                    & continued pre-training & 1603.6 & 1460.9 & --- & --- & --- & $+0.869$ \\
math knowledge                   & continued pre-training & 1876.9 & --- & --- & --- & --- & --- \\
\bottomrule
\end{tabular}}
\end{table}

As shown in Table~\ref{tab:pairs-geometry-full}, we concatenate parameter
differences over backbone tensors (attention, MLP, and normalization
layers) and compute their $\ell_2$ norms and cosine similarities.
The Nemotron measurements additionally include compatible input-embedding
rows after key mapping, excluding protected rows and unmatched parameters.
All measurements exclude output heads, the other model families also
exclude all input embeddings.
We define
$\Delta_b=\bm{D}_b-\bm{A}_b$, $\Delta_p=\bm{D}_p-\bm{A}_p$,
$\bm{\tau}_A=\bm{A}_p-\bm{A}_b$, and $\bm{\tau}_D=\bm{D}_p-\bm{D}_b$.

For most SFT pairs, the AR-to-diffusion conversion direction remains highly stable after post-training, with
$\cos(\Delta_b,\Delta_p)\geq 0.98$, while the AR and diffusion post-training vectors have low cosine similarity, indicating largely distinct updates. DreamReasoner shows a larger change in the conversion direction and also has a substantially larger AR post-training update.

The update scale is also markedly different between post-training and continued pre-training. Post-training vectors have relatively small norms, whereas continued-pretraining vectors are substantially larger, potentially moving the resulting checkpoints beyond the local low-loss region in which weight-space composition is most effective. In addition, the code continued-pretraining vectors are strongly aligned across the AR and diffusion paradigms, in contrast to the near-orthogonality observed for post-training vectors.

\clearpage
\section{Analysis}
\label{app:behav}
\subsection{Correction and error introduction along the A2D direction}

We further characterize how A2D changes model predictions by tracking beneficial and adverse prediction transitions as the AR contribution increases.

\begin{figure}[ht]
\centering
\includegraphics[width=\textwidth]{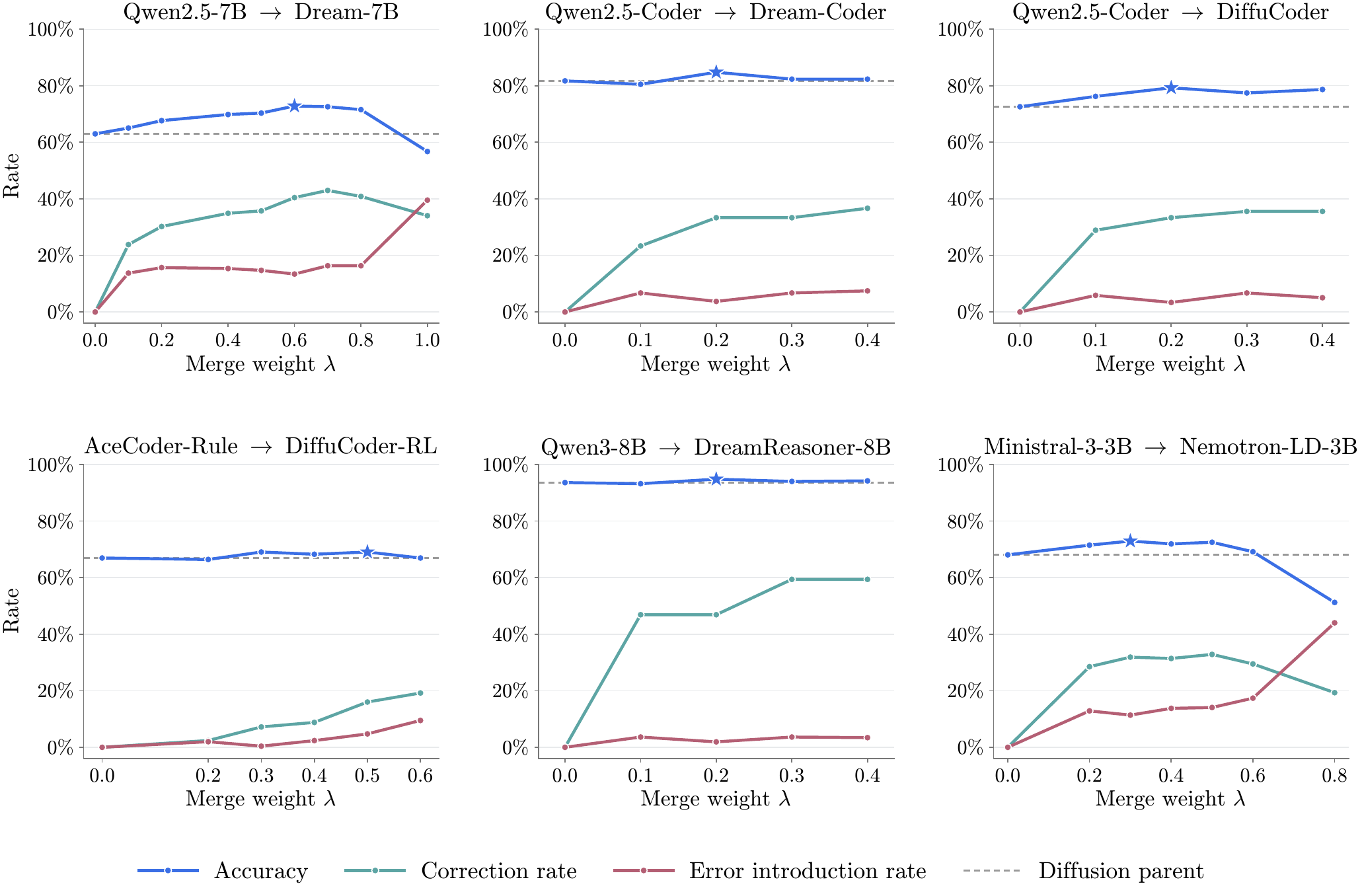}
\caption{\textbf{Correction increases with the AR contribution while error introduction remains limited over the effective interpolation range.}
We report accuracy, correction rate (parent wrong $\to$ A2D correct), and error introduction rate (parent correct $\to$ A2D wrong) along the A2D interpolation path. Dashed lines denote the diffusion parent, and stars mark the best-performing coefficient.}
\label{fig:beta-flips}
\end{figure}

Across the six settings in ~\autoref{fig:beta-flips}, increasing the AR contribution generally increases correction while introducing few new errors around the optimal region. Some models remain robust over a broad range of $\lambda$, while Dream-7B and Nemotron-LD show a clearer trade-off at larger values. Overall, A2D improves performance by correcting parent errors while largely preserving already-correct predictions.

\subsection{Interpolation preserves complementary gains on released checkpoints}

\begin{figure}[ht]
\centering
\includegraphics[width=\textwidth]{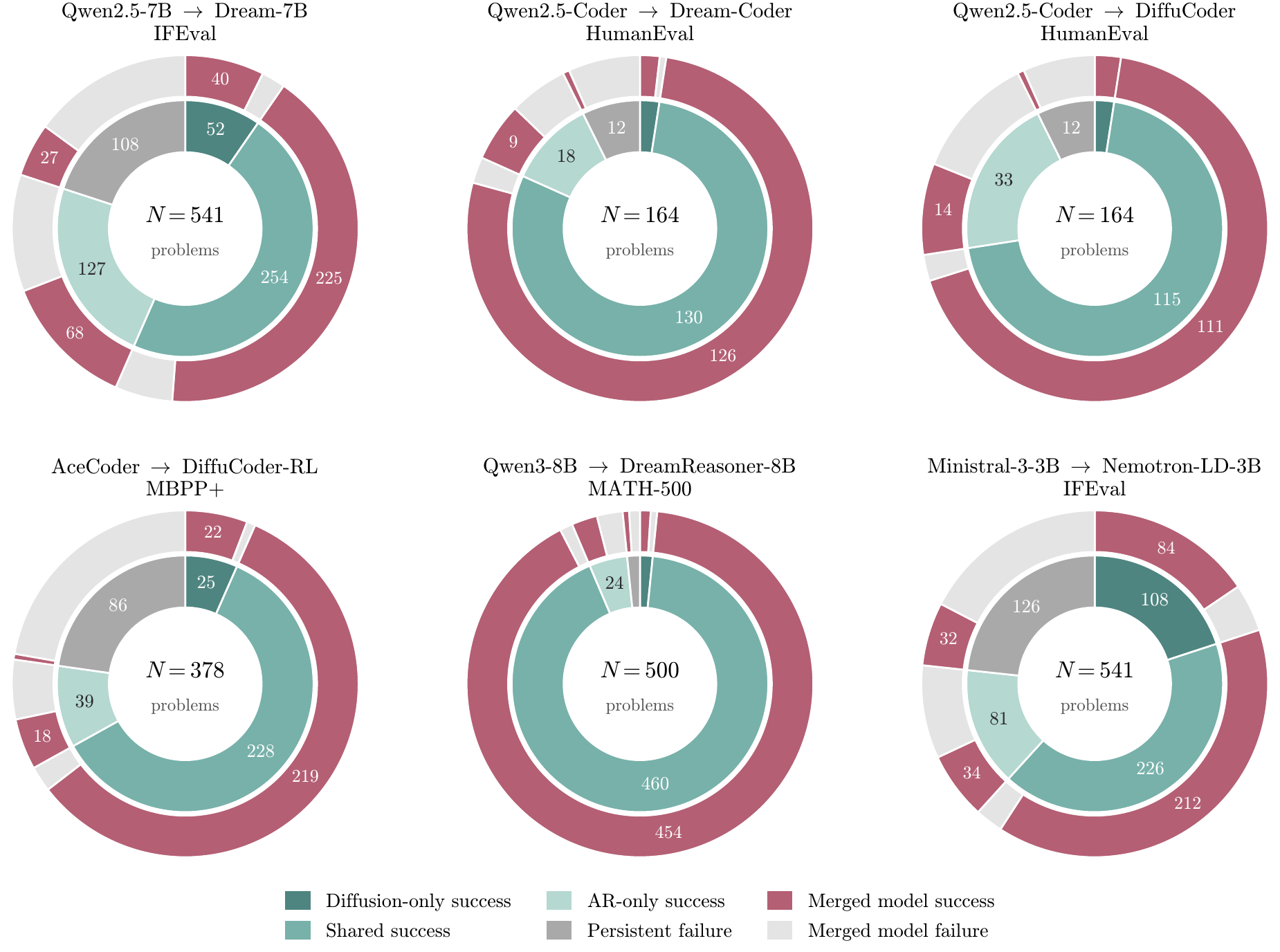}
\caption{\textbf{A2D keeps the successes of both endpoints.} For every item of the benchmark, the inner ring records which of the diffusion parent and the AR donor solves it. The outer ring records
whether the merged model solves it.}
\label{fig:pie_real_ckpts}
\end{figure}

As shown in \autoref{fig:pie_real_ckpts}, the same pattern holds for released checkpoints: the diffusion parent and AR donor solve different questions, while the merged model preserves most successes from both.

\subsection{Aligned representation changes on released checkpoints}

\begin{figure}[ht]
    \centering
    \includegraphics[width=0.6\linewidth]{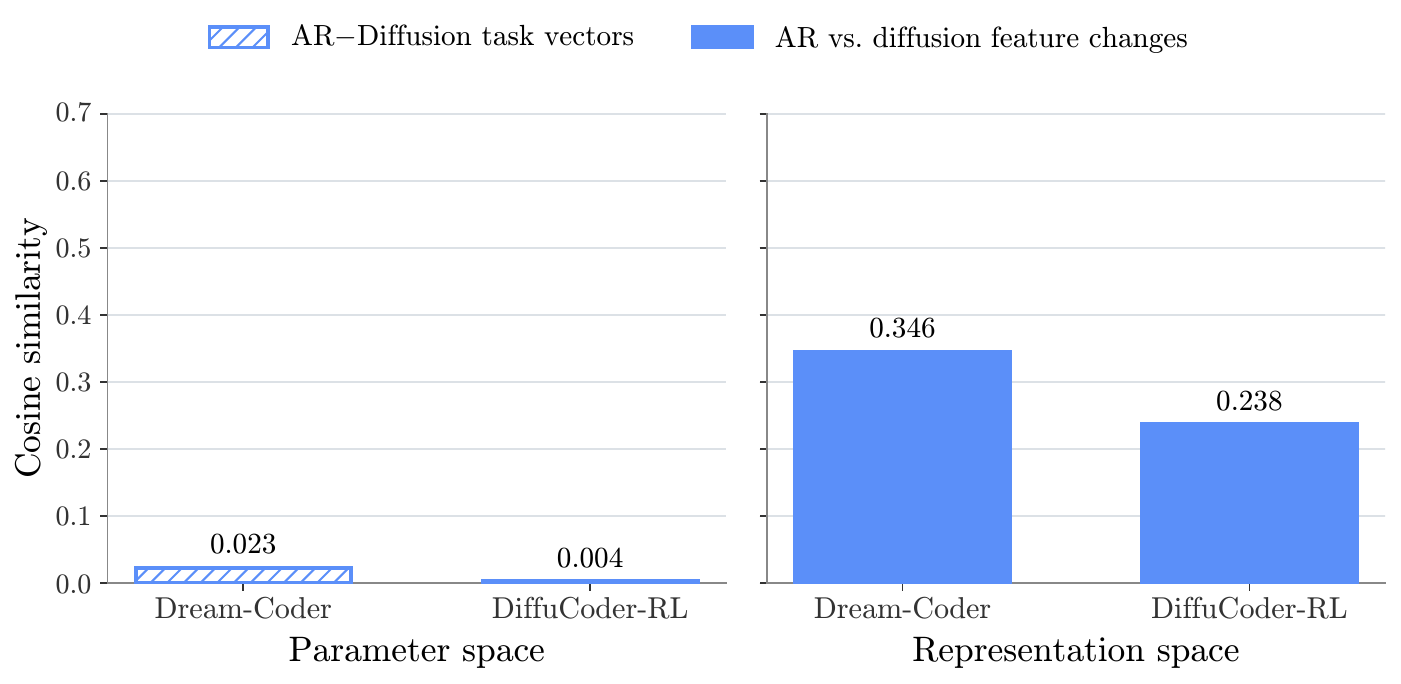}
    \caption{Distinct parameter updates induce aligned representation changes for released checkpoints.}
    \label{fig:rep-align-real}
\end{figure}

The same pattern extends to released checkpoints. As shown in \autoref{fig:rep-align-real}, for both Dream-Coder and DiffuCoder-RL, AR and diffusion post-training updates remain nearly orthogonal in parameter space while inducing substantially more aligned changes in the diffusion model's representations. Notably, this pattern also holds for the RL-derived updates of DiffuCoder-RL.

\clearpage
\section{Limitations and Future Work}
\label{app:disc}

Our results reveal a weight-space connection between AR and diffusion models: AR post-training updates remain meaningful after transfer to diffusion models and can complement diffusion-specific updates. This extends existing connections between the two paradigms beyond training and decoding mechanisms. However, several questions remain open.

A2D currently relies on a compatible training lineage between the AR donor and
diffusion recipient. All model pairs studied here originate from the same AR
family, although our Nemotron experiments show that exact architectural
correspondence is not required. Whether cross-paradigm transfer extends to more
distant or independently trained model pairs remains unclear.

Moreover, the transferability we observe is primarily limited to relatively
local post-training updates. Continued-pretraining updates are substantially
larger and do not transfer effectively in our experiments, suggesting that
pretraining-scale transformations may require additional alignment or
decomposition.

Future work could further characterize what determines cross-paradigm
compatibility, develop more adaptive composition strategies, and study A2D as
an initialization for subsequent SFT or RL.

\end{document}